\pdfoutput=1

\documentclass[11pt, table]{article}

\usepackage[]{acl}

\usepackage{times}
\usepackage{latexsym}

\usepackage[T1]{fontenc}

\usepackage[utf8]{inputenc}

\usepackage{microtype}

\usepackage{graphicx}
\usepackage{multicol, multirow}
\usepackage{booktabs}
\usepackage{float}
\usepackage{amssymb}
\usepackage{amsmath}
\usepackage{enumitem}
\usepackage{hyperref}
\usepackage{bbm}
\usepackage{algorithm, algpseudocode}

\newcommand{\vl}[3]{\parbox[t]{#1}{\multirow{#2}{*}{\rotatebox[origin=c]{90}{#3}}}}
\newcommand{\unws}[1]{\textnormal{\footnotesize\makebox[0pt]{$#1$}}}

\renewcommand{\midrule}{\specialrule{.2pt}{1.5pt}{1.5pt}}

\title{Exploring Description-Augmented Dataless Intent Classification}

\author{Ruoyu Hu$^{1,}$\thanks{$^*$Corresponding author}\ ,
  Foaad Khosmood$^2$ \and Abbas Edalat$^1$\\
  $^1$Department of Computing, Imperial College London, UK \\
  $^2$California Polytechnic State University, USA \\
  \texttt{\{ruoyu.hu18, a.edalat\}@imperial.ac.uk} \\
  \texttt{foaad@calpoly.edu}}

\begin{document}
\maketitle
\begin{abstract}
In this work, we introduce several schemes to leverage description-augmented embedding similarity for dataless intent classification using current state-of-the-art (SOTA) text embedding models. We report results of our methods on four commonly used intent classification datasets and compare against previous works of a similar nature. Our work shows promising results for dataless classification scaling to a large number of unseen intents. We show competitive results and significant improvements (+6.12\% Avg.) over strong zero-shot baselines, all without training on labelled or task-specific data. Furthermore, we provide qualitative error analysis of the shortfalls of this methodology to help guide future research in this area. 
\end{abstract}

\section{Introduction}

Task-oriented dialogue systems (TODS) by design, aid the user in accomplishing tasks within specific domains, and can have a wide range of applications from shopping \cite{yan2017building} to healthcare \cite{wei-etal-2018-task, valizadeh-parde-2022-ai}. Modular TODS \cite{wen-etal-2017-network} will typically contain an intent classification component \cite{louvan-magnini-2020-recent, chen2019bert, su-etal-2022-multi} used by a dialogue manager to determine the appropriate task the user intends to complete. In recent years, neural-based models using supervised training have reached state-of-the-art on many natural language processing tasks, including intent classification. However, supervised learning methods require human-labelled data for a predefined set of intents, which may be time-consuming and labour-intensive to acquire \cite{xia-etal-2018-zero}, and may have poor scalability if new intents are added, or task definition changed. An early approach to tackle this problem is \textit{dataless intent classification} \cite{chang2008importance, song2014dataless} which aimed to leverage the pairwise similarities between semantic representations of utterances and intent classes to perform classification without reliance on human-labelled data. However, this approach relies heavily on the quality of semantic representations \cite{chang2008importance}. In recent years, successful \textit{zero-shot intent classification} approaches \cite{liu-etal-2019-reconstructing, yan-etal-2020-unknown, yin-etal-2019-benchmarking} have received greater attention, whereby learning conducted using labelled examples of a subset of \textit{seen} intent labels is transferred to \textit{unseen} intents. However, these methods still require human-labelled data, and tend to bias towards seen intents, with the number of unseen intents also generally much lower than seen intents \cite{liu2022simple, zhang-etal-2022-learn}. 

In this work, with the significant recent advancements in the quality of text embedding models \cite{muennighoff-etal-2023-mteb}, we explore the potential for dataless intent classification methods using a number of recent state-of-the-art text embedding models. We introduce several approaches for generating intermediate textual representations for intents, most notably using intent label descriptions, and formalise our methodology. We perform extensive evaluation of our methods, including scenarios with large numbers of intents from different domains, using three commonly used intent classification datasets. We summarise our contributions as follows:
\begin{itemize}[noitemsep]
    \item We introduce a new scheme for generating intent descriptions with an aim to minimise reliance on human expert input.
    \item We show that our intent descriptions yield significant improvements over label tokenization through extensive evaluation.
    \item We introduce an approach utilising utterance paraphrasing and masking which yields further improvements and show this is consistent across a range of models.
    \item We aggregate and explore the potential of a multitude of current SOTA text embedding models for dataless classification.
    \item We extensively evaluate our methodology on four commonly used intent classification datasets and report on the results.
    \item We provide qualitative error analysis aimed at guiding future work.
\end{itemize}

\section{Related Works}
\label{sec:related-works}

\subsection{Generalized Zero-Shot Learning}
Zero-shot learning (\textsc{ZSL}) \cite{yin-etal-2019-benchmarking} aims to leverage learning previously performed on labelled examples from seen tasks to unseen tasks, of which there are no labelled examples available for supervised training. \textsc{ZSL} has seen increasing popularity in the domain of intent classification \cite{liu-etal-2019-reconstructing, yan-etal-2020-unknown} in recent years, whereby models are trained on a subset of intent labels and evaluated on another disjoint subset of intent labels. In more recent years, the concept of generalized zero-shot learning (\textsc{GZSL}) has seen an increase in prominence in the domain, in which the performance on both seen and unseen classes are considered in tandem \cite{zhang-etal-2022-learn, lamanov-etal-2022-template}. Several \textsc{GZSL} approaches learn a label prototype space during training, which is transferred to unseen classes through methods such as inter-class relationship modelling \cite{zhang2021generalized} and prototype adaptation \cite{zhang-etal-2022-learn}. Approaches such as \cite{lamanov-etal-2022-template} encode the utterance and labels in a sentence-pair setup, with template-based lexicalisation of labels used as class prototypes. Other approaches exist that use label prototypes as centroids in Gaussian mixture models trained on seen class utterances \cite{yan-etal-2020-unknown, liu2022simple}. An issue that can occur with \textsc{GZSL} is biased towards seen classes \cite{zhang-etal-2022-learn}, which can lead to significantly lower performance on unseen classes. It is also difficult to see the efficacy of transfer to a large number of diverse unseen classes, as the number of unseen classes in evaluation is also typically much smaller than the number of seen classes.

\subsection{Dataless Classification}
Dataless text classification \cite{chang2008importance} is defined as tackling text classification without prior training on any labelled data. Generally regarded as a precursor to zero-shot text classification, this approach typically leverages sentence representations without any training on labelled data, by comparing the semantic representations between a sentence and that of the intent classes \cite{song2014dataless}. \cite{zha2019multilabel} utilises ``seed" words associated with each intent class to further contextualise the intent class representation, as a single word may often be insufficient to encapsulate the meaning of the class \cite{chen2015dataless}. Some approaches further leverage class hierarchy to augment classification performance \cite{li-etal-2016-joint,popov-etal-2019-unsupervised}.

\section{Methodology}
\label{sec:methodology}

\subsection{Problem Definition}
Let $\mathcal{C}$ be a set of intents supported by a task-oriented dialogue system, $\mathcal{U}=\bigcup\{\mathcal{U}_c\}_{c\in\mathcal{C}}$ defines the set of all user utterances, $\mathcal{U}_c=\{u_i\}_{1\leq i \leq n_c}$ is the set of utterances belonging to intent class $c$. The model undergoes no task-specific training and is tasked with making an intent prediction $\hat{y}_i$ for a previously unseen utterance $u_i$ at inference time. We follow the paradigm set by previous works in dataless text classification \cite{chang2008importance, song2014dataless} to conduct nearest-neighbour classification over the sentence embedding space. For a given utterance $u_i$, an encoder $\mathbf{h}(\cdot)$ and a set of class label representations $\{l_c\}_{c\in\mathcal{C}}$, we make a prediction $\hat{y}_i$ as follows:

\begin{equation}
    \hat{y}_i = \underset{c}{\arg\max\ } s(\mathbf{h}(u_i), \mathbf{h}(l_c))
\end{equation}

\noindent where $s(\mathbf{u}, \mathbf{v}) = \mathbf{u}\cdot\mathbf{v} / ||\mathbf{u}||_2||\mathbf{v}||_2$ is the cosine similarity between two vectors.

In order to conduct nearest-neighbour classification using intent labels, we require an intermediate representation, or prototype, which encapsulates to some degree the meaning of a class \cite{zha2019multilabel}, from which we can obtain a suitable embedding. A commonly used approach in dataless classification is to use the labels \cite{chang2008importance}. 

\subsection{Label Tokenization}
\label{sec:methodology:label_tokenization}
A class prototype is obtained by tokenizing intent labels directly, inserting spaces and replacing character separators, i.e.

\begin{table}[ht]
    \centering
    \setlength{\tabcolsep}{1pt}
    \begin{tabular}{ccc}
        \texttt{AddToPlaylist} & $\rightarrow$ & \texttt{Add To Playlist} \\
        \texttt{oil\_change\_how} & $\rightarrow$ & \texttt{Oil Change How}
    \end{tabular}
    \vspace{-5pt}
\end{table}

However, this approach depends on the descriptiveness of the original intent labels, which can vary significantly between datasets and tasks. As such, we propose an additional step to produce intent label \textit{descriptions} which we hypothesise can (1) better align the semantic representation with the characteristics of the class and (2) provide more consistent performance across datasets or approaches without requiring in-task data, which previous works \cite{lamanov-etal-2022-template} have shown could improve performance over purely using tokenized labels.

\subsection{Our Approach}
\subsubsection{Intent Description}

Our objective is to produce a brief description of the intent expressed by the user in a given utterance, while ensuring the process requires minimal expert human effort so as to remain scalable for large numbers of intent classes. Rather than producing a general description of the intent \cite{gao-etal-2023-benefits}, we formalise our template for producing intent descriptions with the two following constraints:

\paragraph{Label Preservation} The resulting intent description must contain tokens from the original intent label i.e. \texttt{car\_rental} $\rightarrow$ \texttt{User wants to rent a car}, or replace with an appropriate word (lexical cognates, synonyms etc.).
\paragraph{Format Consistency} Descriptions should be written in the declarative form, beginning with either \texttt{"User is [asking|saying]"}, or \texttt{"User wants [to]"}, and aim to introduce minimal extraneous tokens in a similar manner to abstractive summarization \cite{de-raedt-etal-2023-idas}. Our approach differs from the template-based approach in \cite{lamanov-etal-2022-template} in that we use exclusively the declarative form in writing our descriptions to maintain consistency across intent classes and datasets. Example descriptions can be seen in Table \ref{tab:example-descriptions}, more examples can be found in Appendix \ref{sec:appendix:intents-desc-examples}. We examine the robustness of our approach in Section \ref{sec:ablations}.

\begin{table}[htb]
    \centering
    \setlength{\tabcolsep}{3pt}
    \begin{tabular}{>{\small}lp{4.8cm}}
        \toprule
        \textbf{Label} & \textbf{Description} \\
        \midrule
        \texttt{abbreviation} & ``user is asking what an abbreviation stands for or means" \\
        \texttt{flight\_no} & ``user is asking about a flight number" \\
        \texttt{AddToPlaylist} & ``user wants to add a song to a playlist" \\
        \texttt{food\_last} & ``user wants to know how long a food lasts \\
        \texttt{maybe} & ``user is expressing uncertainty" \\
        \bottomrule
    \end{tabular}
    \caption{Example descriptions for intent labels from each of the datasets (Section \ref{sec:experiments:datasets}) used in our experimentation.}
    \label{tab:example-descriptions}
    \vspace{-15pt}
\end{table}

In our experimentation (Section \ref{sec:experiments}), our intent descriptions added on average $6.6$ tokens to the tokenized intent labels ($1.9\rightarrow 8.5$), with $98.3\%$ of descriptions containing at least one of the label tokens in exact form, and $82.7\%$ of all label tokens preserved.

\subsubsection{Utterance Paraphrasing}

The diversity of user utterances for any given intent can pose a challenge as intents may not be obvious \cite{mueller-etal-2022-label}. We hypothesise that a format consistency constraint over the user utterance can benefit dataless intent classification performance. Previous works primarily focused on utterance paraphrasing as a means of data augmentation \cite{kumar-etal-2019-submodular, jolly-etal-2020-data, sahu-etal-2022-data} or to reduce overfitting \cite{dopierre-etal-2021-protaugment}. Our approach leverages inference-time paraphrasing to enforce a weaker degree of our intent descriptions' format consistency constraint on user utterances. Given a paraphraser model $\mathbf{p}(\cdot)$ we compute a sentence embedding of the paraphrased utterance $\mathbf{p}(u_i)$:
\begin{equation}
    P_{u_i} = \mathbf{h}(\mathbf{p}(u_i))
\end{equation}
We leverage a 1.6B StableLM model\footnote{\url{https://huggingface.co/stabilityai/stablelm-2-1_6b-chat}} \cite{bellagente2024stable} to generate a single paraphrase for each utterance. Our selection was based on said model being the top-performing model under 2B parameters on the Open LLM Leaderboard \cite{open-llm-leaderboard} as of the time of writing. We additionally experimented with 1.6B Zephyr \cite{tunstall2023zephyr} and 1.3B Phi-1.5 \cite{li2023textbooks} models but found no significant difference on our task. Example templates and further details are shown in Appendix \ref{sec:appendix:paraphrasing}. The mean cosine similarity between the paraphrases and the original utterances across 4 intent classification tasks and 12 embedding models is $0.89\pm 0.06$.

\subsubsection{Label Entity Overlap \& Masking}
\label{sec:methodology:label-entity-overlap}
We note that sentence embeddings tended to capture the topic and entity information rather than the associated action, which can lead to misclassifications in the event that two or more intent classes share entities (i.e. \texttt{AddToPlaylist} and \texttt{PlayMusic} can both refer to songs as their objects). To tackle this, we introduce a masking component which given user utterance $u_i$ masks spans containing the object of said utterance, identified through dependency parsing\footnote{We leverage an off-the-shelf dependency parser, \texttt{en\_core\_web\_trf} from \texttt{Spacy} \textbf{url}: \url{https://spacy.io/models/en}} \cite{de-marneffe-manning-2008-stanford, schuster-manning-2016-enhanced}, to produce $m_i$. $m_i$ is then weighted to form the masking component:
\begin{equation}
    M_{u_i} = \mathbf{h}(m_i) \times Overlaps(u_i, k) \times \mathbbm{1}_{masked}
\end{equation}
where $Overlaps(u, k)$ denotes whether there is likely entity overlap in the top $k$ candidate intents by similarity and $\mathbbm{1}_{masked}$ is whether there exists a masked version of the original sentence. We did not find significant differences in performance for $k>3$, and thus we use $k=3$ for all our experiments.

\paragraph{Masking} Algorithm \ref{algo:appendix:masking} illustrates the masking procedure which identifies and masks object spans in the utterance. We define such object spans as subtrees within the dependency tree in which a parent node has any of \texttt{\{dobj, pobj, ccomp\}} relations. We note that object relations are not always present in the dependency tree, in such cases masked representations are not used. From our experiments, some degree of masking was performed for $97.29\%$ of utterances from the ATIS dataset, $98.04\%$ of SNIPS, $90.88\%$ of CLINC and $84.24\%$ of MASSIVE. We show an example of this procedure in Appendix \ref{sec:appendix:masking-example}.

\begin{algorithm}[t]
    \caption{Utterance Masking Procedure}
    \label{algo:appendix:masking}
    \begin{algorithmic}[1]
        \State Given user utterance $u_i = \{u_{i,1}, \dots u_{i,t}\}$
        \State $T_i$ $\gets$ DependencyParser($u_i$)
        \Procedure{MaskTree}{$T$}
            \State $n$ $\gets$ root($T$)
            \If{relation($n$) is \texttt{obj}}
                \State $n$ $\gets$ \texttt{[MASK]}
                \State DROP children($n$)
            \Else
                \For{$u_{i, j}$ in children($n$)}
                    \State \textsc{MaskTree}($u_{i,j}$)
                \EndFor
            \EndIf
        \EndProcedure
    \end{algorithmic}
\end{algorithm}

\paragraph{Entity Overlap} For each intent, we predict a set of entities $\mathbf{e}_c = \{e_{c_1},\dots,e_{c_k}\}$ from the intent description that may describe the object of said class. As such, entities are defined at problem definition and can be modified alongside intent descriptions when they are added/removed. We precompute an overlap matrix $Overlap$ where
\begin{equation}
    Overlap[i,j] = \begin{cases}
        1 & \mathbf{e}_i \ \cap \mathbf{e}_j \neq \varnothing \\
        0 & \textnormal{otherwise}
    \end{cases}
\end{equation}

At inference time, we compute overlaps for classes with top $k$ embedding similarities for an utterance $u_i$. Given a similarity vector $s_i = \{s_{i,1},\dots,s_{i,c}\}_{c = |\mathcal{C}|}$ of embedding similarities between utterance embedding $\mathbf{h}(u_i)$ and intent description embeddings $\mathbf{h}(l_c)_{c\in\mathcal{C}}$, we compute $Top_k(u_i)$ as the top $k$ classes with similarity scores sorted in descending order. We then compute pairwise overlap for all pairs in $Top_k(u_i)$ as follows:

\begin{equation}
    Overlaps(u_i, k) = \bigcup_\unws{m, n\in Top_k(u_i), m\neq n}\ Overlap[m,n]
\end{equation}

We note that future work could explore expansion of the definition of relevant entities to each intent class, as the current solution relies on the quality of intent descriptions and only covers the most likely entities across an entire class, a more dynamic inference-time solution that determines overlap based on candidate classes would be desirable.

\subsection{Combined Sentence Representation}
We formulate the final representation of the user utterance within the embedding space as the sum of the original utterance embedding with the paraphrasing and masking components:
\begin{align}
    h_i = \mathbf{h}(u_i) + P_{u_i} + M_{u_i} \\
    \hat{y}_i = \underset{c}{\arg\max\ } s(h_i, \mathbf{h}(l_c))
\end{align}

\section{Experiments}
\label{sec:experiments}

\subsection{Datasets}
\label{sec:experiments:datasets}
We evaluate our methods on four commonly used English task-oriented dialogue (TOD) system intent classification datasets, covering a diverse range of number of intents (7-150) and domains (up to 18). (1) \textbf{ATIS} \cite{hemphill1990atis} is an English air-travel information system dataset containing 18 intent classes. For comparison, we follow previous works \cite{zhang-etal-2022-learn} in filtering out intent classes containing fewer than 5 examples. (2) \textbf{SNIPS-NLU} \cite{coucke2018snips} contains 7 intent classes, totalling 14,484 utterances. (3) \textbf{CLINC} \cite{larson-etal-2019-evaluation} is a dataset for out-of-scope intent classification, with 150 intents and 22,500 utterances spanning 10 domains. (4) \textbf{MASSIVE} \cite{fitzgerald-etal-2023-massive} is a multilingual spoken language understanding dataset containing 60 intents across 18 domains, we select the 16,521 instances from the \texttt{en-US} split of the dataset for our experiments. As our method does not involve fine-tuning on task-specific data, we consider \textit{entire} datasets to consist of unseen data for evaluation\footnote{We make our code and datasets publicly available and can be found at \url{https://github.com/ruoyunlp/dataless-intent-classification}}.

\subsection{Models}

We select 11 models from the Massive Text Embedding Benchmark (MTEB) \cite{muennighoff-etal-2023-mteb} that are in the top 20 at the time of writing\footnote{November-December 2023. We note our top-performing selected models are still competitive with current top-performing models from MTEB fitting our criteria as of May 2024}. Our selections are based on the following criteria: (1) the model weights must be released (2) documentation of training methods and experimentation details must be readily available. Additionally, owing to computational limits\footnote{All experiments conducted using a single ~9GB GPU}, we only consider models up to 3GB in size. Our final selection of 11 models can be largely grouped into 4 families of models: \textbf{InstructOR} \cite{su-etal-2023-one}, \textbf{E5} \cite{wang2022text}, \textbf{GTE} \cite{li2023general} and \textbf{BGE} \cite{xiao2023cpack}. More details on selected models are provided in Appendix \ref{sec:appendix:models}.

We report results in Section \ref{sec:results} for all E5, GTE and BGE models using averaged token embeddings as sentence representations. We additionally compare model performances against a commonly used embedding model in OpenAI's \texttt{text-embedding-ada-002} \cite{neelakantan2022text} which we refer to in our tables as `Ada-002'. We also investigated the generation of synthetic examples as intent prototypes (Appendix \ref{sec:appendix:intents-desc-examples}) but did not find significant improvements over our approach using intent descriptions (Appendix \ref{sec:appendix:synthetic-full}).

\section{Results}
\label{sec:results}

\subsection{Baselines and Terminology}
We compare the performance of our methods against several unknown intent classification methods previously detailed in Section \ref{sec:related-works}. Here we clarify the terminology used henceforth to refer to these methods in our results. We refer to scores on unseen intent labels reported by \cite{zhang2021generalized} as \textbf{ICR}, \cite{yan-etal-2020-unknown} as \textbf{SEG}, \cite{liu2022simple} as \textbf{ML-SEG}, dataless approach trained using original data from \cite{lamanov-etal-2022-template} as \textbf{TIR}$_{Orig}$ and likewise \textbf{TIR}$_{Syn}$ for training on synthetic data. We refer to the results of the adapted method of \cite{gidaris2018dynamic} reported in \cite{zhang-etal-2022-learn} as \textbf{CosT} and the reported main results as \textbf{LTA}. We refer to the best-performing model of a similar size to our selection from \cite{gretz-etal-2023-zero} as \textbf{TTC}$_{D}$.

\subsection{Metrics}
Following from previous works \cite{zhang-etal-2022-learn, lamanov-etal-2022-template}, we report Accuracy and Macro-F1 scores for intent classification on each of the datasets, in addition, we also compute the average of Accuracy and F1 score for direct comparison between our methods similar to \cite{gritta-etal-2022-crossaligner}. We show macro-F1 only for MASSIVE in Table \ref{tab:main-results} for comparison's sake as the previous work \cite{gretz-etal-2023-zero} did not report Accuracy scores. Full results for each of our approaches including Accuracy scores are shown in Table \ref{tab:appendix:main-results-expanded}.

\begin{table}[!ht]
    \normalsize
	\centering
	\setlength{\tabcolsep}{1pt}
     \bgroup
     \def\arraystretch{0.95}
    \begin{tabular}{c|l||c|c|c|c||c}
         \toprule
         & \multirow{2}{*}{\textbf{Model}} & \textbf{AT.} & \textbf{SN.} & \textbf{CL.} & \textbf{MA.} & \multirow{2}{*}{\textbf{Ovr.}}\\
         \cline{3-6}
         & & \multicolumn{3}{c|}{Mean Acc. \& F1} & F1 & \\
        \midrule
\vl{1em}{8}{\textit{Baselines}}  & ICR & \cellcolor[HTML]{fba877} 35.04 & - & - & - & -\\
 & SEG & - & \cellcolor[HTML]{fba677} 69.46 & - & - & -\\
 & ML-SEG & - & \cellcolor[HTML]{fdc47d} 76.53 & - & - & -\\
 & TIR$_{Orig}$ & - & - & \cellcolor[HTML]{fdc47d} 68.50 & - & -\\
 & TIR$_{Syn}$ & - & - & \cellcolor[HTML]{f97c6f} 59.65 & - & -\\
 & CosT & \cellcolor[HTML]{fed17f} 45.62 & \cellcolor[HTML]{f8696b} 55.28 & \cellcolor[HTML]{fcb479} 66.50 & - & -\\
 & LTA  & \cellcolor[HTML]{93cc7e} 60.55 & \cellcolor[HTML]{dce182} 87.16 & \cellcolor[HTML]{e8e483} 74.46 & - & -\\
 & TTC$_{D}$ & - & - & - & \cellcolor[HTML]{fed981} 54.22 & -\\
 \midrule
 & \textit{Baselines} & \cellcolor[HTML]{93cc7e} 60.55 & \cellcolor[HTML]{dce182} 87.16 & \cellcolor[HTML]{e8e483} 74.46 & \cellcolor[HTML]{fed981} 54.22 & \cellcolor[HTML]{dae082} 69.10\\
\midrule
\vl{1em}{12}{\textit{Tokenized Intent Labels}}  & Instr.$_{Large}$ & \cellcolor[HTML]{f8696b} 18.72 & \cellcolor[HTML]{fedd81} 82.39 & \cellcolor[HTML]{fa9574} 62.76 & \cellcolor[HTML]{fa8e72} 47.62 & \cellcolor[HTML]{fa9473} 52.87\\
 & E5-v2$_{Base}$ & \cellcolor[HTML]{f86f6c} 20.39 & \cellcolor[HTML]{fdc77d} 77.13 & \cellcolor[HTML]{fb9e75} 63.87 & \cellcolor[HTML]{f97c6f} 45.97 & \cellcolor[HTML]{fa8d72} 51.84\\
 & E5-v2$_{Large}$ & \cellcolor[HTML]{fa8871} 26.64 & \cellcolor[HTML]{fba877} 69.99 & \cellcolor[HTML]{f98270} 60.40 & \cellcolor[HTML]{fa8570} 46.83 & \cellcolor[HTML]{fa8871} 50.97\\
 & mE5$_{Large}$ & \cellcolor[HTML]{f9786e} 22.47 & \cellcolor[HTML]{f97a6e} 59.35 & \cellcolor[HTML]{f8696b} 57.34 & \cellcolor[HTML]{f8696b} 44.34 & \cellcolor[HTML]{f8696b} 45.88\\
 & E5$_{Large}$ & \cellcolor[HTML]{fdbe7b} 40.57 & \cellcolor[HTML]{fcbb7b} 74.44 & \cellcolor[HTML]{fdc97d} 69.11 & \cellcolor[HTML]{fba777} 49.78 & \cellcolor[HTML]{fcb67a} 58.48\\
 & Ada-002 & \cellcolor[HTML]{fa8570} 25.98 & \cellcolor[HTML]{fedf82} 82.75 & \cellcolor[HTML]{fcb87a} 66.97 & \cellcolor[HTML]{fa9173} 47.90 & \cellcolor[HTML]{fba677} 55.90\\
 & GTE$_{Small}$ & \cellcolor[HTML]{f8716d} 20.75 & \cellcolor[HTML]{fcb97a} 73.99 & \cellcolor[HTML]{fdc47c} 68.47 & \cellcolor[HTML]{fdbf7c} 51.90 & \cellcolor[HTML]{fb9974} 53.77\\
 & GTE$_{Base}$ & \cellcolor[HTML]{d3de81} 55.66 & \cellcolor[HTML]{feda81} 81.75 & \cellcolor[HTML]{fed680} 70.65 & \cellcolor[HTML]{fcba7b} 51.44 & \cellcolor[HTML]{fedd81} 64.88\\
 & GTE$_{Large}$ & \cellcolor[HTML]{fcbb7b} 39.78 & \cellcolor[HTML]{fed07f} 79.36 & \cellcolor[HTML]{fdcd7e} 69.54 & \cellcolor[HTML]{fb9f75} 49.08 & \cellcolor[HTML]{fcbc7b} 59.44\\
 & BGE$_{Small}$ & \cellcolor[HTML]{f86c6c} 19.50 & \cellcolor[HTML]{fdca7e} 78.00 & \cellcolor[HTML]{fed780} 70.78 & \cellcolor[HTML]{fdc57d} 52.43 & \cellcolor[HTML]{fba276} 55.18\\
 & BGE$_{Base}$ & \cellcolor[HTML]{fed27f} 45.74 & \cellcolor[HTML]{fdc57d} 76.81 & \cellcolor[HTML]{ffe984} 73.05 & \cellcolor[HTML]{fdea84} 55.89 & \cellcolor[HTML]{fed17f} 62.87\\
 & BGE$_{Large}$ & \cellcolor[HTML]{fdcc7e} 44.17 & \cellcolor[HTML]{fed27f} 79.67 & \cellcolor[HTML]{ffeb84} 73.25 & \cellcolor[HTML]{fedd81} 54.53 & \cellcolor[HTML]{fed17f} 62.91\\
\midrule
\vl{1em}{12}{\textit{Intent Label Descriptions}}  & Instr.$_{Large}$ & \cellcolor[HTML]{fdc47c} 42.18 & \cellcolor[HTML]{ffeb84} 85.60 & \cellcolor[HTML]{b4d580} 77.25 & \cellcolor[HTML]{ffe883} 55.52 & \cellcolor[HTML]{fedf82} 65.14\\
 & E5-v2$_{Base}$ & \cellcolor[HTML]{fdea84} 52.44 & \cellcolor[HTML]{d5df82} 87.49 & \cellcolor[HTML]{fed880} 70.92 & \cellcolor[HTML]{fed480} 53.73 & \cellcolor[HTML]{ffe583} 66.14\\
 & E5-v2$_{Large}$ & \cellcolor[HTML]{ffea84} 52.16 & \cellcolor[HTML]{d9e082} 87.31 & \cellcolor[HTML]{fedd81} 71.49 & \cellcolor[HTML]{ffea84} 55.65 & \cellcolor[HTML]{ffe883} 66.65\\
 & mE5$_{Large}$ & \cellcolor[HTML]{93cc7e} 60.51 & \cellcolor[HTML]{ffe483} 83.88 & \cellcolor[HTML]{ffe382} 72.24 & \cellcolor[HTML]{f1e783} 56.67 & \cellcolor[HTML]{e9e583} 68.32\\
 & E5$_{Large}$ & \cellcolor[HTML]{fcea84} 52.56 & \cellcolor[HTML]{b5d680} 88.92 & \cellcolor[HTML]{e0e282} 74.88 & \cellcolor[HTML]{f6e984} 56.32 & \cellcolor[HTML]{ece583} 68.17\\
 & Ada-002 & \cellcolor[HTML]{ffe783} 51.34 & \cellcolor[HTML]{a8d27f} 89.50 & \cellcolor[HTML]{a9d27f} 77.81 & \cellcolor[HTML]{dce182} 58.03 & \cellcolor[HTML]{d8e082} 69.17\\
 & GTE$_{Small}$ & \cellcolor[HTML]{dfe282} 54.71 & \cellcolor[HTML]{ffe683} 84.42 & \cellcolor[HTML]{fed27f} 70.20 & \cellcolor[HTML]{fdbf7b} 51.86 & \cellcolor[HTML]{fee082} 65.30\\
 & GTE$_{Base}$ & \cellcolor[HTML]{fbea84} 52.60 & \cellcolor[HTML]{ede683} 86.41 & \cellcolor[HTML]{dce182} 75.10 & \cellcolor[HTML]{fede81} 54.62 & \cellcolor[HTML]{ffeb84} 67.18\\
 & GTE$_{Large}$ & \cellcolor[HTML]{d0de81} 55.85 & \cellcolor[HTML]{efe683} 86.33 & \cellcolor[HTML]{cedd81} 75.83 & \cellcolor[HTML]{dfe282} 57.85 & \cellcolor[HTML]{dce182} 68.97\\
 & BGE$_{Small}$ & \cellcolor[HTML]{feda81} 47.84 & \cellcolor[HTML]{ffeb84} 85.51 & \cellcolor[HTML]{fee182} 72.03 & \cellcolor[HTML]{feda81} 54.27 & \cellcolor[HTML]{fedd81} 64.91\\
 & BGE$_{Base}$ & \cellcolor[HTML]{fedd81} 48.76 & \cellcolor[HTML]{c2d980} 88.32 & \cellcolor[HTML]{add37f} 77.61 & \cellcolor[HTML]{cedd81} 58.92 & \cellcolor[HTML]{e7e483} 68.40\\
 & BGE$_{Large}$ & \cellcolor[HTML]{dde182} 54.88 & \cellcolor[HTML]{acd37f} 89.30 & \cellcolor[HTML]{91cb7e} 79.08 & \cellcolor[HTML]{90cb7e} \underline{62.88} & \cellcolor[HTML]{abd37f} 71.53\\
\midrule
\vl{1em}{12}{\textit{+ Paraphrase and Masking}}  & Instr.$_{Large}$ & \cellcolor[HTML]{fede82} 49.07 & \cellcolor[HTML]{a0cf7e} 89.86 & \cellcolor[HTML]{7cc57c} \underline{80.17} & \cellcolor[HTML]{c0d980} 59.79 & \cellcolor[HTML]{cedd81} 69.72\\
 & E5-v2$_{Base}$ & \cellcolor[HTML]{8eca7d} 60.93 & \cellcolor[HTML]{9cce7e} 90.03 & \cellcolor[HTML]{dde182} 75.06 & \cellcolor[HTML]{dfe282} 57.81 & \cellcolor[HTML]{b6d680} 70.95\\
 & E5-v2$_{Large}$ & \cellcolor[HTML]{fedb81} 48.06 & \cellcolor[HTML]{ffeb84} 85.56 & \cellcolor[HTML]{e4e382} 74.69 & \cellcolor[HTML]{d8e082} 58.27 & \cellcolor[HTML]{ffe883} 66.64\\
 & mE5$_{Large}$ & \cellcolor[HTML]{b8d780} 57.72 & \cellcolor[HTML]{fee182} 83.36 & \cellcolor[HTML]{dee182} 75.00 & \cellcolor[HTML]{e1e282} 57.67 & \cellcolor[HTML]{e7e483} 68.43\\
 & E5$_{Large}$ & \cellcolor[HTML]{ece583} 53.78 & \cellcolor[HTML]{72c27c} \underline{91.92} & \cellcolor[HTML]{c6db81} 76.27 & \cellcolor[HTML]{cadc81} 59.17 & \cellcolor[HTML]{c3da81} 70.28\\
 & Ada-002 & \cellcolor[HTML]{c1d980} 57.02 & \cellcolor[HTML]{91cb7e} 90.51 & \cellcolor[HTML]{85c87d} 79.73 & \cellcolor[HTML]{bed880} 59.92 & \cellcolor[HTML]{a5d17f} 71.80\\
 & GTE$_{Small}$ & \cellcolor[HTML]{f0e783} 53.48 & \cellcolor[HTML]{c7db81} 88.11 & \cellcolor[HTML]{fedd81} 71.50 & \cellcolor[HTML]{e4e382} 57.53 & \cellcolor[HTML]{f6e883} 67.66\\
 & GTE$_{Base}$ & \cellcolor[HTML]{63be7b} \textbf{64.20} & \cellcolor[HTML]{f9e984} 85.88 & \cellcolor[HTML]{d0dd81} 75.75 & \cellcolor[HTML]{d6df82} 58.41 & \cellcolor[HTML]{b4d580} 71.06\\
 & GTE$_{Large}$ & \cellcolor[HTML]{92cb7e} 60.63 & \cellcolor[HTML]{76c47c} 91.70 & \cellcolor[HTML]{95cc7e} 78.89 & \cellcolor[HTML]{a3d17f} 61.63 & \cellcolor[HTML]{8ac97d} \underline{73.21}\\
 & BGE$_{Small}$ & \cellcolor[HTML]{e7e483} 54.16 & \cellcolor[HTML]{8cca7d} 90.76 & \cellcolor[HTML]{dde182} 75.04 & \cellcolor[HTML]{cbdc81} 59.11 & \cellcolor[HTML]{cddd81} 69.77\\
 & BGE$_{Base}$ & \cellcolor[HTML]{abd37f} 58.69 & \cellcolor[HTML]{74c37c} 91.81 & \cellcolor[HTML]{83c77d} 79.80 & \cellcolor[HTML]{9ecf7e} 61.98 & \cellcolor[HTML]{8dca7d} 73.07\\
 & BGE$_{Large}$ & \cellcolor[HTML]{8cca7d} \underline{61.04} & \cellcolor[HTML]{63be7b} \textbf{92.57} & \cellcolor[HTML]{63be7b} \textbf{81.52} & \cellcolor[HTML]{63be7b} \textbf{65.76} & \cellcolor[HTML]{63be7b} \textbf{75.22}\\
        \bottomrule
    \end{tabular}
    \egroup
    \vspace{-5pt}
    \caption{Model performance on 4 intent classification tasks. We show Mean of Accuracy and Macro-F1 scores for ATIS, SNIPS-NLU \& CLINC. Macro-F1 is shown for MASSIVE as TTC$_D$ did not report Accuracy. Full results for each dataset are shown in Table \ref{tab:appendix:main-results-expanded}.}
    \label{tab:main-results}
    \vspace{-15pt}
\end{table}

\begin{table}[t]
    \centering
    \bgroup
    \def\arraystretch{0.95}
    \begin{tabular}{l|c|c|c}
        \toprule
        \textbf{Model} & \textbf{Tok.} & \textbf{Desc.} & \textbf{Comb.} \\
        \midrule
InstructOR$_{Large}$ & \cellcolor[HTML]{fcb67a} 64.96 & \cellcolor[HTML]{f0e783} 73.19 & \cellcolor[HTML]{a2d07f} 76.89\\
E5-v2$_{Base}$ & \cellcolor[HTML]{fba777} 62.98 & \cellcolor[HTML]{fee182} 71.02 & \cellcolor[HTML]{d3de81} 74.58\\
E5-v2$_{Large}$ & \cellcolor[HTML]{fa9073} 59.75 & \cellcolor[HTML]{ffe683} 71.76 & \cellcolor[HTML]{f1e783} 73.13\\
mE5$_{Large}$ & \cellcolor[HTML]{f8696b} 54.23 & \cellcolor[HTML]{ffe483} 71.50 & \cellcolor[HTML]{fdea84} 72.57\\
E5$_{Large}$ & \cellcolor[HTML]{fcb479} 64.70 & \cellcolor[HTML]{e6e483} 73.65 & \cellcolor[HTML]{b3d580} 76.09\\
Ada-002 & \cellcolor[HTML]{fdc07c} 66.48 & \cellcolor[HTML]{c2da81} 75.35 & \cellcolor[HTML]{9dcf7e} 77.12\\
GTE$_{Small}$ & \cellcolor[HTML]{fcb97a} 65.43 & \cellcolor[HTML]{fed580} 69.38 & \cellcolor[HTML]{f8e984} 72.80\\
GTE$_{Base}$ & \cellcolor[HTML]{fecf7f} 68.57 & \cellcolor[HTML]{ffea84} 72.35 & \cellcolor[HTML]{e7e483} 73.63\\
GTE$_{Large}$ & \cellcolor[HTML]{fdc17c} 66.63 & \cellcolor[HTML]{e8e483} 73.57 & \cellcolor[HTML]{94cc7e} 77.57\\
BGE$_{Small}$ & \cellcolor[HTML]{fdcd7e} 68.20 & \cellcolor[HTML]{fee182} 71.11 & \cellcolor[HTML]{c2d980} 75.37\\
BGE$_{Base}$ & \cellcolor[HTML]{fed580} 69.36 & \cellcolor[HTML]{c4da81} 75.28 & \cellcolor[HTML]{8ac97d} \underline{78.05}\\
BGE$_{Large}$ & \cellcolor[HTML]{fed880} 69.76 & \cellcolor[HTML]{9dcf7e} 77.15 & \cellcolor[HTML]{63be7b} \textbf{79.91}\\
        \bottomrule
    \end{tabular}
    \egroup
    \caption{Average model Mean of Accuracy and F1 over SNIPS-NLU, CLINC and MASSIVE datasets using tokenized intent labels (\textbf{Tok.}), intent descriptions (\textbf{Desc.}) and combined utterance embedding (\textbf{Comb.}).}
    \label{tab:overall-scores}
    \vspace{-15pt}
\end{table}

\subsection{Methods using Tokenized Labels} Despite a lack of task-specific fine-tuning, models using tokenized intent labels generally performed comparably to most of the baselines on unseen intents. The best-performing model (BGE$_{Large}$) outperforms baseline scores for ICR ($+9.13$ Mean), SEG (+10.21 Mean) and ML-SEG (+3.14 Mean), TIR$_{Syn}$ (+13.60 Mean), TIR$_{Orig}$ (+4.55 Mean) and TTC$_D$ (+0.31 F1). BGE$_{Large}$ outperforms CosT on all datasets; however, it also significantly underperforms LTA on all 3 datasets (-16.38 ATIS, -7.49 SNIPS-NLU, -1.21 CLINC). We note that this approach appears quite sensitive to the model as indicated by the comparatively high standard deviation ($\sigma_{Ovr}=5.65$) across models.


\subsection{Methods using Intent Descriptions}

Our method using intent label descriptions yields a significant improvement over using tokenized labels (Tables \ref{tab:main-results} and \ref{tab:overall-scores}), with an average increase per model of $+11.24$ overall. This supports our hypothesis (1) (Section \ref{sec:methodology:label_tokenization}) in that the additional contextualisation added through describing the label via a declarative sentence better encapsulates the semantic information represented by a label. We also note from Table \ref{tab:overall-scores} that the standard deviation in performance across models is significantly lower when using descriptions ($\sigma_{Ovr}=1.98$), supporting our hypothesis (2) that descriptions can improve consistency across models and approaches. Our overall best-performing model (BGE$_{Large}$) also considerably outperforms the strongest baseline on SNIPS-NLU (+2.14 Mean), CLINC (+4.62 Mean) and MASSIVE (+8.66 F1). We do note that all of our approaches in this setup underperform on the ATIS dataset compared to the baseline, with our overall best-performing approach yielding 60.51 vs 60.55; we provide further insight into possible reasons in Section \ref{sec:analysis} to help guide future research.

\subsection{Methods with Additional Paraphrasing and Masking}
Our addition of paraphrase and masked utterance embeddings yields further overall score improvements on average of +3.16 over label descriptions and is consistent across different models (Table \ref{tab:overall-scores}). Our best-performing model (BGE$_{Large}$) significantly outperforms previous approaches on all 4 datasets (+0.49 ATIS, +5.42 SNIPS-NLU, +7.06 CLINC, +11.54 MASSIVE). Additionally, our approach outperforms previous work on 9 out of 12 selected models.

\section{Ablations}
\label{sec:ablations}

\paragraph{Addition of paraphrasing and masking} Table \ref{tab:overall-scores} illustrates the mean performance across SNIPS, CLINC and MASSIVE datasets for each model different class prototypes. We note the consistent improvement in performance between tokenized intent labels and our approach using declarative intent descriptions (+7.86 Mean), and the further improvements with added paraphrasing and masking (+10.56 Mean). We omit ATIS from this table as it is significantly unbalanced, the impact of which we explore in Section \ref{sec:analysis}, and its results are already included in Table \ref{tab:main-results}.

\paragraph{Combination of techniques} Table \ref{tab:ablations} demonstrates the performance (mean of accuracy and macro-f1) between different combinations of our techniques using a \texttt{bge-large-en-v1.5} model. We observe that the addition of paraphrasing increases performance by an average of +2.06\% compared to methods without, supporting our hypothesis (3) that inference-time paraphrasing can benefit dataless intent classification. We observe that masking increases performance by an average of +1.80\% and the addition of masked embedding only when entity overlaps are predicted increases performance by +0.32\% on average. We perform further ablations over combinations of techniques using other models in Appendix \ref{sec:appendix:further-ablations} and note similar behaviour across different models.

\begin{table}[t]
    \centering
    \setlength{\tabcolsep}{2pt}
    \bgroup
    \def\arraystretch{0.95}
    \begin{tabular}{c|c|c|c||c|c|c|c||c}
        \toprule
        \multicolumn{4}{l||}{\textbf{Setup}} & \multirow{2}{*}{\textbf{AT.}} & \multirow{2}{*}{\textbf{SN.}} & \multirow{2}{*}{\textbf{CL.}} & \multirow{2}{*}{\textbf{MA.}} & \multirow{2}{*}{\textbf{Ovr.}} \\
        \cline{0-3}\textbf{E} & \textbf{P} & \textbf{M} & \textbf{O} & & & & \\
        \midrule
    x & & & & 54.89 & 89.29 & 79.08 & 63.09 & \cellcolor[HTML]{ffe583} 71.59\\
     & x & & & 56.03 & 85.77 & 78.77 & 63.35 & \cellcolor[HTML]{ffe282} 70.98\\
     & & x & & 30.72 & 76.76 & 37.90 & 33.62 & \cellcolor[HTML]{f8696b} 44.75\\
     \midrule
     x & x & & & 56.11 & 88.83 & \textbf{81.56} & 65.60 & \cellcolor[HTML]{fcea84} 73.02\\
     x & & x & & 60.84 & 92.52 & 75.56 & 60.80 & \cellcolor[HTML]{ffe984} 72.43\\
     & x & x & & 60.57 & 92.19 & 75.99 & 62.91 & \cellcolor[HTML]{ffeb84} 72.92\\
     x & x & x & & \textbf{61.04} & \textbf{92.67} & 81.22 & \underline{65.64} & \cellcolor[HTML]{67bf7b} \underline{75.14}\\
     \midrule
     x & & x & x & 60.84 & 92.56 & 77.36 & 61.82 & \cellcolor[HTML]{f3e883} 73.14\\
     & x & x & x & 60.57 & 92.02 & 76.86 & 63.04 & \cellcolor[HTML]{f5e883} 73.12\\
     x & x & x & x & \textbf{61.04} & \underline{92.57} & \underline{81.52} & \textbf{65.65} & \cellcolor[HTML]{63be7b} \textbf{75.20}\\
        \bottomrule
    \end{tabular}
    \egroup
    \caption{Mean of Accuracy and Macro-F1 on 4 intent classification datasets using a \texttt{bge-large-en-v1.5} model. \textbf{Setup} denotes whether a component is used in the combined sentence embedding: \textbf{E} - utterance embedding, \textbf{P} - paraphrasing, \textbf{M} - masking, \textbf{O} - entity overlap in masking.}
    \label{tab:ablations}
    \vspace{-15pt}
\end{table}

\begin{table*}[!h]
    \centering
    \setlength{\tabcolsep}{2pt}
    \begin{tabular}{l||c|c|c|c||c}
    \toprule
    \textbf{Setup} & \textbf{ATIS} & \textbf{SNIPS} & \textbf{CLINC} & \textbf{MASSIVE} & \textbf{Overall} \\
    \midrule
    Tokenized Intent Labels & 40.11 & 78.74 & 72.45 & 54.53 & 61.46 \\
    \midrule
Intent Label Descriptions & $42.00 \pm 3.91$ & $86.97 \pm 2.05$ & $73.77 \pm 1.10$ & $61.12 \pm 1.04$ & $65.97 \pm 2.02$ \\ 
+ Paraphrase \& Masking & $\textbf{46.83} \pm 4.18$ & $\textbf{91.21} \pm 1.61$ & $\textbf{76.17} \pm 1.14$ & $\textbf{63.61} \pm 1.19$ & $\textbf{69.46} \pm 2.03$ \\
    \bottomrule
    \end{tabular}
    \caption{Comparison of macro-f1 score across 200 sampled combinations of descriptions for our setups with/without paraphrasing and masking. Note our combined approach outperforms tokenized labels across all datasets.}
    \label{tab:ablations-desc-paraphrases}
\end{table*}

\begin{figure*}[!h]
    \centering
    \begin{minipage}{0.48\linewidth}
        \centering
        \includegraphics[width=\linewidth]{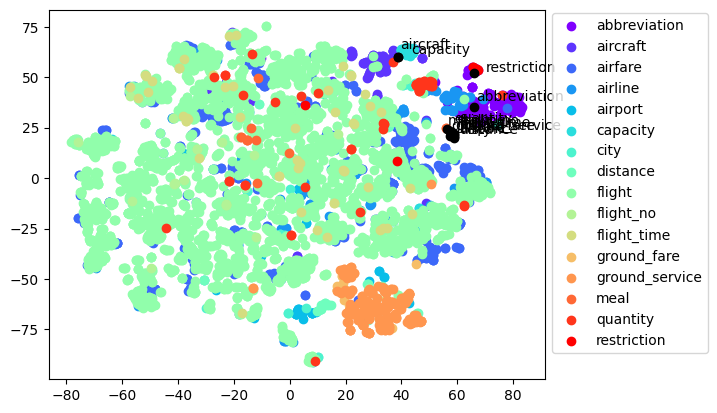}
        \vspace{-25pt}
        \caption*{(\textbf{a})}
    \end{minipage}
    \begin{minipage}{0.50\linewidth}
        \centering
        \includegraphics[width=\linewidth]{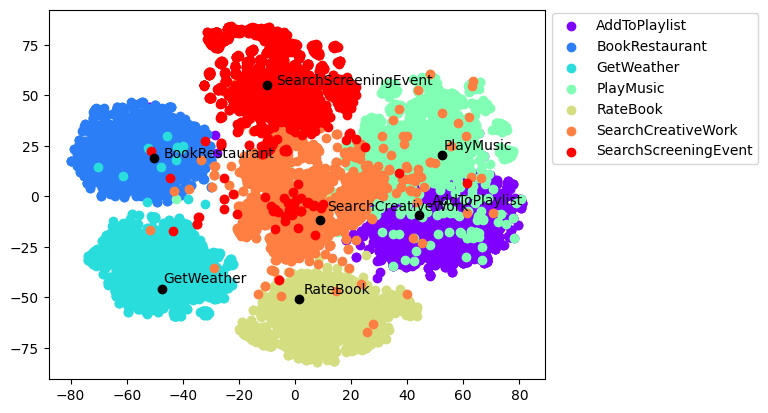}
        \vspace{-25pt}
        \caption*{(\textbf{b})}
    \end{minipage}
    \begin{minipage}{0.48\linewidth}
        \centering
        \includegraphics[width=\linewidth]{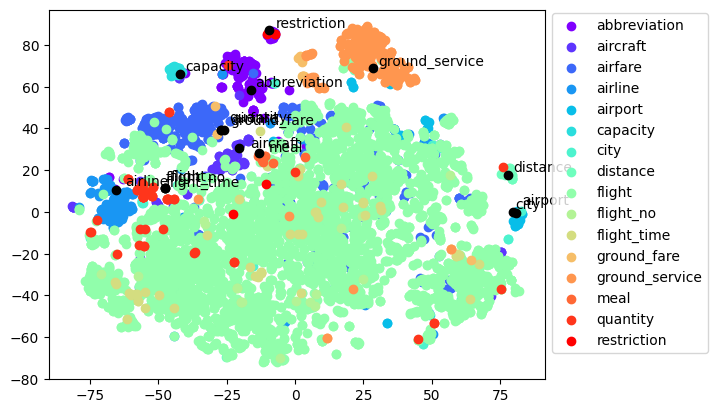}
        \vspace{-25pt}
        \caption*{(\textbf{c})}
    \end{minipage}
    \begin{minipage}{0.50\linewidth}
        \centering
        \includegraphics[width=\linewidth]{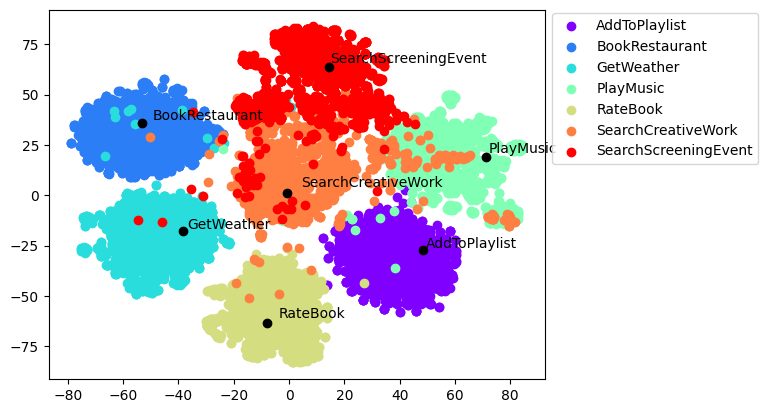}
        \vspace{-25pt}
        \caption*{(\textbf{d})}
    \end{minipage}
    \vspace{-10pt}
    \caption{t-SNE \cite{vandermaaten2008visualizing} visualisation of embeddings computed using BGE$_{Large}$, class label description embeddings are shown in black and labelled. (\textbf{Row 1}) Embeddings of ATIS (\textbf{a}) and SNIPS (\textbf{b}), (\textbf{Row 2}) Embeddings with Paraphrasing and Masking for ATIS (\textbf{c}) and SNIPS (\textbf{d}).}
    \label{fig:embedding-visualisation}
    \vspace{-15pt}
\end{figure*}

\paragraph{Choice of Descriptions} To investigate whether our proposed method is sensitive to the choice of intent descriptions, we generate paraphrases of our manually produced descriptions with increasing temperature values and sampled 200 combinations of descriptions for each dataset. Table \ref{tab:ablations-desc-paraphrases} contains the mean and standard deviations of the macro-f1 scores for each dataset, we report macro-f1 for this ablations experiment due to the severely unbalanced nature of the ATIS dataset towards a single class \texttt{flight} (accounting for $\sim 74\%$ of the dataset). Further details on description paraphrase generation and sampling along with examples are provided in Appendix \ref{sec:appendix:ablations-desc-paraphrases}. Methods using only tokenized intent labels are outperformed by our methods using label descriptions ($+4.51\%$), with further improvements from the addition of paraphrasing and masking ($+8.00\%$). The overall scores per dataset are slightly affected by the choice of intent descriptions, with standard deviations between 1-2\% with the exception of the ATIS dataset. Future work could focus on the combination of multiple intent descriptions (via paraphrasing) or description refinement with unsupervised training \cite{chu-etal-2021-unsupervised, muller-etal-2022-shot} to further improve robustness to the choice of descriptions.

\section{Analysis and Future Work}
\label{sec:analysis}

\paragraph{In-Domain Saturation} We visualise the embeddings generated by our best-performing model (BGE$_{Large}$) on the 4 evaluation datasets using t-SNE \cite{vandermaaten2008visualizing}, along with the embedding for the intent label description to gain insight into the source of errors in our approach. Figure \ref{fig:embedding-visualisation} shows the distribution of embeddings on the ATIS and SNIPS datasets. In the interest of space, visualisations of CLINC and MASSIVE are shown in Appendix \ref{sec:appendix:tsne-visualisation} . We observe a poor alignment on the ATIS dataset between the intent label descriptions (Figure \ref{fig:embedding-visualisation}a) and utterance embeddings corresponding to each class, possibly explaining the poor performance in general on this dataset across models. We note the single-domain nature of the ATIS dataset, with all utterances relating to air-travel/flight; additionally, we note the significantly imbalanced nature of the ATIS dataset \cite{nan-etal-2021-uncovering}, with $\sim 74\%$ of utterances belonging to the \texttt{flight} class, which is a label that overlaps the domain of the dataset. We hypothesise this may lead to the intent label descriptions being much worse at capturing semantic information distinct to each class. This is supported by analysis of the pairwise embedding similarities of utterances belonging to the same class vs utterances belonging to different classes (Table \ref{tab:mean-similarity-dataset}) where models' embeddings on the ATIS dataset consistently had lower percentage-difference in embedding similarity between \textit{in}-class and \textit{out}-class, implying more difficulty in distinguishing the utterances using solely embeddings. This issue is mitigated to some degree with our addition of paraphrasing and masking, as the number of misclassifications where there are entity overlaps between classes is reduced on average by $19.19\%$. We see this visually in Figure \ref{fig:embedding-visualisation}d as the cluster for each class is more distinct compared to \ref{fig:embedding-visualisation}b. Errors from classes with overlapping entities in SNIPS are reduced by $29.31\%$.

\paragraph{Error Analysis}
We perform qualitative analysis of the remaining errors and identify two categories of commonly occurring errors. (1) \textit{Description Scope:} Our approach utilises a single description for each intent and can work well when an intent concerns a limited number of topics; however, intents such as \texttt{meta} and \texttt{small\_talk} from the CLINC dataset, and \texttt{qa} from the MASSIVE dataset can encompass a significantly broader range of topics than other intents within the same dataset. The impact of topical granularity per intent class and the potential for a hierarchical approach to intent classes in a dataless setting can be the focus of future work in this area. (2) \textit{Action Overlap:} Our approach mitigates some errors arising from shared entities across intents through masking. Whilst this has shown success in reducing errors of this nature (i.e. between \texttt{PlayMusic} and \texttt{AddToPlaylist} from the SNIPS dataset), it is less successful in events where an action is shared across classes, such as \texttt{play} from the MASSIVE dataset, and  \texttt{SearchCreativeWork} and \texttt{SearchScreeningEvent} from the SNIPS-NLU dataset. Future work could investigate the potential to decouple the desired \textit{action} and \textit{object} (topical information) in utterance embeddings.

\paragraph{Label Candidate Analysis} We observed from our results (Table \ref{tab:main-results}) that our approach, despite outperforming strong baselines on ATIS and MASSIVE datasets, still consistently underperforms compared to the same setup on SNIPS-NLU and CLINC. We therefore investigate the position of the correct label when ranking embedding similarities. Table \ref{tab:label-top-k} shows the percentage of examples where the correct label is ranked within the top-\textit{k} by embedding similarity for $k=1,3,5,10$. We note for erroneous predictions, the correct label is within the Top-3 in $67.11\%$ of cases, $81.89\%$ in Top-5 and $90.94\%$ in Top-10. This implies that our approach can be used to identify candidate intents from a larger set of intents, with a high success rate even for small values of $k$ (i.e. 91.01\% Top-3).

\begin{table}[t]
    \centering
    \setlength{\tabcolsep}{4pt}
    \bgroup
    \def\arraystretch{0.95}
    \begin{tabular}{l|c|c|c|c}
        \toprule
        \textbf{Dataset} & \textbf{Top-1} & \textbf{Top-3} & \textbf{Top-5} & \textbf{Top-10} \\
        \midrule
ATIS & 67.70 & 93.38 & 96.03 & 98.10 \\
SNIPS-NLU & 89.78 & 97.13 & 99.43 & 100.00 \\
CLINC & 77.24 & 91.71 & 94.86 & 97.41 \\
MASSIVE & 61.45 & 81.85 & 87.79 & 92.79 \\
\midrule
Average & 74.04 & 91.01 & 94.53 & 97.08 \\
        \bottomrule
    \end{tabular}
    \egroup
    \vspace{-5pt}
    \caption{Percentage of correct labels within Top-\textit{k} ranked by embedding similarity per evaluation dataset, averaged across 11 selected models.}
    \label{tab:label-top-k}
    \vspace{-15pt}
\end{table}

\paragraph{Analysis Summary} Our proposed approach performs well overall against the strong baseline methods in unseen intent classification; however, it struggles in certain instances with overlaps in intents within the same domain. We identified potential areas for future work to pursue in tackling said issues. The results of our experiments have shown intent label descriptions can perform well as intent prototypes in this problem setting, and that the addition of paraphrasing and masking can further improve performance.

\paragraph{Limitations} This approach contains a number of limitations: We have identified issues with the descriptiveness of individual labels earlier in this section, and textual labels may not be readily available for certain datasets, though summarisation methods may be effectively applied to a few user utterances to produce such labels. Our evaluation compares against previous works using scores as reported in their respective papers, further work can be done to replicate their experiments to mitigate any potential risk arising from differences in experimental settings. Future work may also investigate the application of descriptions to tasks outside of intent classification, such as emotion recognition \cite{rashkin-etal-2019-towards}.

\section{Conclusion}
Dataless classification allows for scaling to a large number of unseen classes without requiring training on labelled, task-specific data. The benefits of such an approach can enhance development of task-oriented dialogue systems in application to data-poor or compute-limited scenarios where supported intents may also change as the system is developed. In this paper, we have explored the potential of current SOTA text embedding models in dataless intent classification settings using three different approaches for representing intent classes and compared our results against strong zero-shot learning baselines. We proposed a method for standardising the generation of intent label descriptions with an aim to minimise the amount of human annotation required to further support scaling to high numbers of intent classes. Our results have shown that description-augmented dataless classification methods can achieve comparable, and sometimes superior performance to zero-shot methods on the task of intent classification.

\section{Acknowledgements}
We would like to thank Anandha Gopalan for his helpful comments on the paper. Student Ruoyu Hu was funded by UKRI CDT in AI4Health - grant number EP/S023283/1.

\bibliography{anthology,custom}
\bibliographystyle{acl_natbib}

\appendix

\section{Utterance Paraphrasal}
\label{sec:appendix:paraphrasing}

Table \ref{tab:appendix:paraphrase-template} contains an example template used to generate paraphrases for utterances from the CLINC dataset. Examples used in the template do not appear in the dataset and do not make explicit mentions of classes. We use \texttt{length\_penalty}=-1 to encourage shorter outputs, \texttt{repetition\_penalty}=1.2 and \texttt{num\_beams}=3, we use default values for all other generation parameters.

\begin{table}[h]
    \centering
    \small
    \begin{tabular}{p{0.9\linewidth}}
        \toprule
        \textbf{Prompt} \\
        \midrule
Given an utterance, describe what the user is asking. \\ \\
sentence: "set an alarm for every weekday at 7 am" \\
description: user is asking to set an alarm for every weekday at 7am \\ \\
sentence: "can you show me the step-by-step instructions to bake chocolate chip cookies" \\
description: user is asking for recipe for chocolate chip cookies \\ \\
sentence: "could you please tell me what time it is now" \\
description: user is asking for the current time \\ \\
sentence: "\{\}" \\
description:
    \end{tabular}
    \caption{Example template used to generate user utterance paraphrases from the CLINC dataset.}
    \label{tab:appendix:paraphrase-template}
\end{table}

We perform an additional ablation study over the choice of examples in the paraphrase generation template using 9 different examples across 3 configurations for each of SNIPS and MASSIVE datasets. We select these datasets specifically as we believe they differ sufficiently in number of intents and domains. Across 3 ablation configurations and the original paraphrasing setup, we obtain an overall score (mean of accuracy and macro-f1) of $92.66\pm 0.19$\% for SNIPS and $65.48\pm 0.18$\% for MASSIVE. As the standard deviation is low in both instances, we conclude that the choice of examples in the paraphrase generation prompt has little impact on the final performance through our setup.

\section{Example Masking Procedure}
\label{sec:appendix:masking-example}

\begin{figure*}[h]
    \centering
    \includegraphics[width=0.9\linewidth]{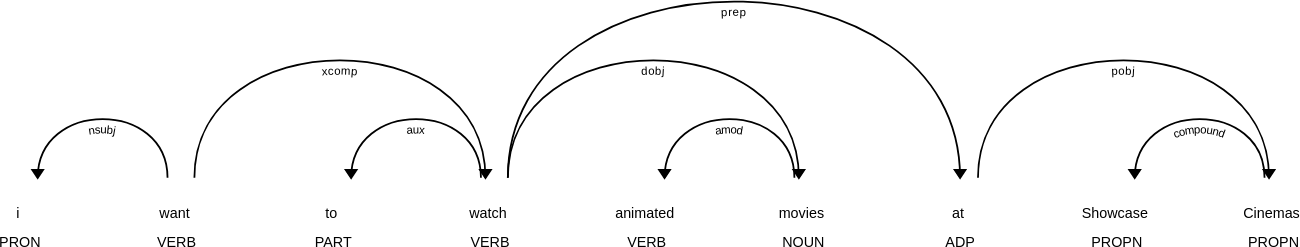}
    \caption{Example dependency parse tree from the SNIPS dataset.}
    \label{fig:appendix:masking-example}
\end{figure*}

Given an user utterance ``i want to watch animated movies at Showcase Cinemas'', we first perform dependency parsing to identify utterance objects that can be masked. Figure \ref{fig:appendix:masking-example} shows an illustration of the resulting parsed dependency relations. Following the approach outlined in Section \ref{sec:methodology:label-entity-overlap}, we mask out nodes with any of \texttt{\{dobj, pobj, ccomp\}} relations, namely ``animated movies'' and ``Showcase Cinemas'' to produce the resulting masked representation ``i want to watch [MASK] at [MASK]''.

\section{Details of selected models}
\label{sec:appendix:models}
Basic model specifications are shown in Table \ref{tab:model-spec}.

\begin{table}[ht]
    \centering
    \setlength{\tabcolsep}{3pt}
    \begin{tabular}{l|cccc}
        \toprule
        \textbf{Model} & $s$ & $d_h$ & $l$ & $\mu_{\mathbf{MTEB}}$ \\
        \midrule
        InstructOR$_{Large}$ & 1.34 & 768 & 512 & 61.59 \\
        \midrule
        E5-v2$_{Base}$ & 0.44 & 768 & 512 & 61.50 \\
        E5-v2$_{Large}$ & 1.34 & 1024 & 512 & 62.25 \\
        Multilingual-E5$_{Large}$ & 2.24 & 1024 & 514 & 61.50 \\
        E5$_{Large}$ & 1.34 & 1024 & 512 & 61.42 \\
        \midrule
        GTE$_{Small}$ & 0.07 & 384 & 512 & 61.36 \\
        GTE$_{Base}$ & 0.22 & 768 & 512 & 62.39 \\
        GTE$_{Large}$ & 0.67 & 1024 & 512 & 63.13 \\
        \midrule
        BGE$_{Small}$ & 0.13 & 384 & 512 & 62.17 \\
        BGE$_{Base}$ & 0.44 & 768 & 512 & 63.55 \\
        BGE$_{Large}$ & 1.34 & 1024 & 512 & 64.23 \\
        \midrule
        OpenAI-Ada-002 & - & 1536 & 8191 & 60.99 \\
        \bottomrule
    \end{tabular}
    \caption{Specifications of selected models grouped by training method. Column $s$ shows model size (GB), $d_h$ embedding dimensions, $l$ maximum sequence length and $\mu_{\mathbf{MTEB}}$ averaged performance on MTEB benchmark.}
    \label{tab:model-spec}
    \vspace{-15pt}
\end{table}

\paragraph{InstructOR} \cite{su-etal-2023-one} embeds the utterance with a task description, allowing for task-specific conditioning at inference time, with good performance on unseen domains. Trained on 330 datasets using a contrastive learning objective \cite{ni-etal-2022-large}. This family of models is initialised from GTR \cite{ni-etal-2022-large} models, which are in-turn initialised from T5 \cite{raffel2020exploring} models.

\paragraph{E5} \cite{wang2022text} performs unsupervised pretraining on the model on $\sim$270M text pairs using an InfoNCE \cite{oord2019representation} objective with other utterances within the batch acting as negative examples, followed by supervised fine-tuning on 3 datasets. We select the \textit{Base} and \textit{Large} variants, initialised from \textit{bert-base-uncased} and \textit{bert-large-uncased-whole-word-masking} respectively.

\paragraph{GTE} \cite{li2023general} pretrains the model on $\sim$800M text pairs and fine-tunes using 33 datasets. The contrastive learning objective used in this work considers, for each query-document pair $(q_i, d_i)$ in a batch, the pairwise relation to the remaining examples $\{(q_j, d_j)\}_{j\neq i}$. The embedding similarities  $s(q_i, d_j)$, $s(q_i, q_j)$, $s(d_i, d_j)$ are added to the partition function, where $s(q,d)$ is the cosine similarity between two embeddings.

\paragraph{BGE} The work~\cite{xiao2023cpack} initialised from BERT \cite{devlin-etal-2019-bert} models and trained using RetroMAE \cite{xiao-etal-2022-retromae} whereby both the input sentence and sentence embeddings in an autoencoder setup are randomly masked during MLM training. The authors use \texttt{[CLS]} token embeddings as the sentence representation. Our experimentation showed a slight improvement when using averaged token embeddings (Mean performance $+0.82\%$ \textit{Tokenized-labels}, $+1.06\%$ \textit{Class-description}).

\section{Full Results}
\label{sec:appendix:full-results-expanded}
See Table \ref{tab:appendix:main-results-expanded} for individual accuracy and macro-f1 scores by task and model.

\begin{table*}
    \centering
    \setlength{\tabcolsep}{1pt}
    \begin{tabular}{c|l||c|c|c||c|c|c||c|c|c||c|c|c}
        \toprule
        & \multirow{2}{*}{\textbf{Model}} & \multicolumn{3}{c||}{\textbf{ATIS}} & \multicolumn{3}{c||}{\textbf{SNIPS}} & \multicolumn{3}{c||}{\textbf{CLINC}} & \multicolumn{3}{c}{\textbf{MASSIVE}}\\
        \cline{3-14}
        & & Acc & F1 & Mean & Acc & F1 & Mean & Acc & F1 & Mean & Acc & F1 & Mean \\
        \midrule
\vl{1em}{8}{\textit{Baselines}}  & ICR & \cellcolor[HTML]{fba777} 35.54 & \cellcolor[HTML]{fcb179} 34.54 & \cellcolor[HTML]{fba877} 35.04 & - & - & - & - & - & - & - & - & -\\
 & SEG & - & - & - & \cellcolor[HTML]{fcb279} 69.61 & \cellcolor[HTML]{fb9f75} 69.31 & \cellcolor[HTML]{fba677} 69.46 & - & - & - & - & - & -\\
 & ML-SEG & - & - & - & \cellcolor[HTML]{fdcb7e} 77.08 & \cellcolor[HTML]{fdc17c} 75.97 & \cellcolor[HTML]{fdc47d} 76.53 & - & - & - & - & - & -\\
 & TIR$_{Orig}$ & - & - & - & - & - & - & \cellcolor[HTML]{fb9c75} 63.90 & \cellcolor[HTML]{f9e984} 73.10 & \cellcolor[HTML]{fdc47d} 68.50 & - & - & -\\
 & TIR$_{Syn}$ & - & - & - & - & - & - & \cellcolor[HTML]{f8696b} 58.00 & \cellcolor[HTML]{fb9874} 61.30 & \cellcolor[HTML]{f97c6f} 59.65 & - & - & -\\
 & CosT & \cellcolor[HTML]{fdc27c} 46.04 & \cellcolor[HTML]{dbe182} 45.21 & \cellcolor[HTML]{fed17f} 45.62 & \cellcolor[HTML]{f8696b} 47.73 & \cellcolor[HTML]{f97e6f} 62.84 & \cellcolor[HTML]{f8696b} 55.28 & \cellcolor[HTML]{fa9273} 62.73 & \cellcolor[HTML]{fed981} 70.28 & \cellcolor[HTML]{fcb479} 66.50 & - & - & -\\
 & LTA  & \cellcolor[HTML]{dde182} 66.09 & \cellcolor[HTML]{63be7b} \textbf{55.02} & \cellcolor[HTML]{93cc7e} 60.55 & \cellcolor[HTML]{a7d27f} 90.09 & \cellcolor[HTML]{ffeb84} 84.22 & \cellcolor[HTML]{dce182} 87.16 & \cellcolor[HTML]{ffeb84} 73.18 & \cellcolor[HTML]{c7db81} 75.74 & \cellcolor[HTML]{e8e483} 74.46 & - & - & -\\
 & TTC$_{D}$ & - & - & - & - & - & - & - & \cellcolor[HTML]{f8696b} 54.73 & - & - & \cellcolor[HTML]{fed981} 54.22 & -\\
 \midrule
 & \textit{Baselines} & \cellcolor[HTML]{dde182} 66.09 & \cellcolor[HTML]{63be7b} \textbf{55.02} & \cellcolor[HTML]{93cc7e} 60.55 & \cellcolor[HTML]{a7d27f} 90.09 & \cellcolor[HTML]{ffeb84} 84.22 & \cellcolor[HTML]{dce182} 87.16 & \cellcolor[HTML]{ffeb84} 73.18 & \cellcolor[HTML]{c7db81} 75.74 & \cellcolor[HTML]{e8e483} 74.46 & - & \cellcolor[HTML]{fed981} 54.22 & -\\
\midrule
\vl{1em}{12}{\textit{Tokenized Intent Labels}}  & Instr.$_{Large}$ & \cellcolor[HTML]{f86c6b} 12.41 & \cellcolor[HTML]{f8696b} 25.03 & \cellcolor[HTML]{f8696b} 18.72 & \cellcolor[HTML]{fede81} 82.71 & \cellcolor[HTML]{fee082} 82.07 & \cellcolor[HTML]{fedd81} 82.39 & \cellcolor[HTML]{fba176} 64.50 & \cellcolor[HTML]{fa9674} 61.02 & \cellcolor[HTML]{fa9574} 62.76 & \cellcolor[HTML]{fb9874} 51.86 & \cellcolor[HTML]{fa8e72} 47.62 & \cellcolor[HTML]{fa9373} 49.74\\
 & E5-v2$_{Base}$ & \cellcolor[HTML]{f86e6c} 13.20 & \cellcolor[HTML]{f97c6f} 27.58 & \cellcolor[HTML]{f86f6c} 20.39 & \cellcolor[HTML]{fdcc7e} 77.30 & \cellcolor[HTML]{fdc67d} 76.96 & \cellcolor[HTML]{fdc77d} 77.13 & \cellcolor[HTML]{fba877} 65.33 & \cellcolor[HTML]{fba076} 62.40 & \cellcolor[HTML]{fb9e75} 63.87 & \cellcolor[HTML]{f98370} 49.91 & \cellcolor[HTML]{f97c6f} 45.97 & \cellcolor[HTML]{f97f6f} 47.94\\
 & E5-v2$_{Large}$ & \cellcolor[HTML]{f8716d} 14.67 & \cellcolor[HTML]{fecf7f} 38.61 & \cellcolor[HTML]{fa8871} 26.64 & \cellcolor[HTML]{fcb67a} 70.83 & \cellcolor[HTML]{fb9e75} 69.15 & \cellcolor[HTML]{fba877} 69.99 & \cellcolor[HTML]{fa8771} 61.56 & \cellcolor[HTML]{fa8971} 59.24 & \cellcolor[HTML]{f98270} 60.40 & \cellcolor[HTML]{fa8d72} 50.88 & \cellcolor[HTML]{fa8570} 46.83 & \cellcolor[HTML]{fa8971} 48.85\\
 & mE5$_{Large}$ & \cellcolor[HTML]{f9766d} 16.41 & \cellcolor[HTML]{f98370} 28.53 & \cellcolor[HTML]{f9786e} 22.47 & \cellcolor[HTML]{fa9273} 59.90 & \cellcolor[HTML]{f8696b} 58.80 & \cellcolor[HTML]{f97a6e} 59.35 & \cellcolor[HTML]{f9736d} 59.13 & \cellcolor[HTML]{f86f6c} 55.56 & \cellcolor[HTML]{f8696b} 57.34 & \cellcolor[HTML]{f8696b} 47.63 & \cellcolor[HTML]{f8696b} 44.34 & \cellcolor[HTML]{f8696b} 45.98\\
 & E5$_{Large}$ & \cellcolor[HTML]{fdbe7b} 44.71 & \cellcolor[HTML]{fdbf7c} 36.43 & \cellcolor[HTML]{fdbe7b} 40.57 & \cellcolor[HTML]{fdc67d} 75.68 & \cellcolor[HTML]{fcb379} 73.21 & \cellcolor[HTML]{fcbb7b} 74.44 & \cellcolor[HTML]{fed27f} 70.27 & \cellcolor[HTML]{fdc87d} 67.96 & \cellcolor[HTML]{fdc97d} 69.11 & \cellcolor[HTML]{fa9273} 51.30 & \cellcolor[HTML]{fba777} 49.78 & \cellcolor[HTML]{fb9c75} 50.54\\
 & Ada-002 & \cellcolor[HTML]{f98470} 21.88 & \cellcolor[HTML]{fa8f72} 30.09 & \cellcolor[HTML]{fa8570} 25.98 & \cellcolor[HTML]{fee082} 83.32 & \cellcolor[HTML]{fee182} 82.19 & \cellcolor[HTML]{fedf82} 82.75 & \cellcolor[HTML]{fdc17c} 68.25 & \cellcolor[HTML]{fcb87a} 65.70 & \cellcolor[HTML]{fcb87a} 66.97 & \cellcolor[HTML]{fa9473} 51.50 & \cellcolor[HTML]{fa9173} 47.90 & \cellcolor[HTML]{fa9273} 49.70\\
 & GTE$_{Small}$ & \cellcolor[HTML]{f8706c} 14.28 & \cellcolor[HTML]{f9796e} 27.21 & \cellcolor[HTML]{f8716d} 20.75 & \cellcolor[HTML]{fdc47d} 74.94 & \cellcolor[HTML]{fcb279} 73.04 & \cellcolor[HTML]{fcb97a} 73.99 & \cellcolor[HTML]{fdca7e} 69.38 & \cellcolor[HTML]{fdc57d} 67.55 & \cellcolor[HTML]{fdc47c} 68.47 & \cellcolor[HTML]{fdc47d} 55.78 & \cellcolor[HTML]{fdbf7c} 51.90 & \cellcolor[HTML]{fdc07c} 53.84\\
 & GTE$_{Base}$ & \cellcolor[HTML]{c5da81} 68.99 & \cellcolor[HTML]{feeb84} 42.34 & \cellcolor[HTML]{d3de81} 55.66 & \cellcolor[HTML]{fedd81} 82.37 & \cellcolor[HTML]{fedb81} 81.14 & \cellcolor[HTML]{feda81} 81.75 & \cellcolor[HTML]{fedd81} 71.56 & \cellcolor[HTML]{fed580} 69.74 & \cellcolor[HTML]{fed680} 70.65 & \cellcolor[HTML]{fdbd7b} 55.15 & \cellcolor[HTML]{fcba7b} 51.44 & \cellcolor[HTML]{fcba7b} 53.30\\
 & GTE$_{Large}$ & \cellcolor[HTML]{fdc07c} 45.14 & \cellcolor[HTML]{fcb079} 34.42 & \cellcolor[HTML]{fcbb7b} 39.78 & \cellcolor[HTML]{fed580} 80.13 & \cellcolor[HTML]{fdce7e} 78.60 & \cellcolor[HTML]{fed07f} 79.36 & \cellcolor[HTML]{fed47f} 70.44 & \cellcolor[HTML]{fdcd7e} 68.64 & \cellcolor[HTML]{fdcd7e} 69.54 & \cellcolor[HTML]{fba476} 52.88 & \cellcolor[HTML]{fb9f75} 49.08 & \cellcolor[HTML]{fba176} 50.98\\
 & BGE$_{Small}$ & \cellcolor[HTML]{f8696b} 11.40 & \cellcolor[HTML]{f97c6f} 27.60 & \cellcolor[HTML]{f86c6c} 19.50 & \cellcolor[HTML]{fed27f} 79.20 & \cellcolor[HTML]{fdc57d} 76.81 & \cellcolor[HTML]{fdca7e} 78.00 & \cellcolor[HTML]{fede82} 71.67 & \cellcolor[HTML]{fed680} 69.89 & \cellcolor[HTML]{fed780} 70.78 & \cellcolor[HTML]{ffeb84} 59.21 & \cellcolor[HTML]{fdc57d} 52.43 & \cellcolor[HTML]{fed680} 55.82\\
 & BGE$_{Base}$ & \cellcolor[HTML]{fed17f} 52.15 & \cellcolor[HTML]{fed580} 39.34 & \cellcolor[HTML]{fed27f} 45.74 & \cellcolor[HTML]{fdcd7e} 77.73 & \cellcolor[HTML]{fdc07c} 75.88 & \cellcolor[HTML]{fdc57d} 76.81 & \cellcolor[HTML]{f3e883} 73.85 & \cellcolor[HTML]{ffe783} 72.24 & \cellcolor[HTML]{ffe984} 73.05 & \cellcolor[HTML]{dee282} 60.55 & \cellcolor[HTML]{fdea84} 55.89 & \cellcolor[HTML]{f5e883} 58.22\\
 & BGE$_{Large}$ & \cellcolor[HTML]{fdc77d} 48.24 & \cellcolor[HTML]{fedb81} 40.11 & \cellcolor[HTML]{fdcc7e} 44.17 & \cellcolor[HTML]{fed780} 80.60 & \cellcolor[HTML]{fdcf7f} 78.74 & \cellcolor[HTML]{fed27f} 79.67 & \cellcolor[HTML]{f0e783} 74.05 & \cellcolor[HTML]{ffe984} 72.45 & \cellcolor[HTML]{ffeb84} 73.25 & \cellcolor[HTML]{fedf82} 58.19 & \cellcolor[HTML]{fedd81} 54.53 & \cellcolor[HTML]{fedc81} 56.36\\
\midrule
\vl{1em}{12}{\textit{Intent Label Descriptions}}  & Instr.$_{Large}$ & \cellcolor[HTML]{fcb67a} 41.24 & \cellcolor[HTML]{f5e883} 43.12 & \cellcolor[HTML]{fdc47c} 42.18 & \cellcolor[HTML]{ffe884} 85.85 & \cellcolor[HTML]{eae583} 85.35 & \cellcolor[HTML]{ffeb84} 85.60 & \cellcolor[HTML]{aad37f} 77.95 & \cellcolor[HTML]{b8d780} 76.55 & \cellcolor[HTML]{b4d580} 77.25 & \cellcolor[HTML]{fedd81} 57.95 & \cellcolor[HTML]{ffe883} 55.52 & \cellcolor[HTML]{fee082} 56.73\\
 & E5-v2$_{Base}$ & \cellcolor[HTML]{e8e483} 64.84 & \cellcolor[HTML]{feda81} 40.04 & \cellcolor[HTML]{fdea84} 52.44 & \cellcolor[HTML]{e2e382} 87.75 & \cellcolor[HTML]{c5da81} 87.23 & \cellcolor[HTML]{d5df82} 87.49 & \cellcolor[HTML]{ffe282} 72.15 & \cellcolor[HTML]{fed580} 69.68 & \cellcolor[HTML]{fed880} 70.92 & \cellcolor[HTML]{fdc27c} 55.57 & \cellcolor[HTML]{fed480} 53.73 & \cellcolor[HTML]{fdc97e} 54.65\\
 & E5-v2$_{Large}$ & \cellcolor[HTML]{fdea84} 62.33 & \cellcolor[HTML]{ffe984} 41.98 & \cellcolor[HTML]{ffea84} 52.16 & \cellcolor[HTML]{e0e282} 87.84 & \cellcolor[HTML]{cedd81} 86.77 & \cellcolor[HTML]{d9e082} 87.31 & \cellcolor[HTML]{ffe483} 72.39 & \cellcolor[HTML]{fedb81} 70.59 & \cellcolor[HTML]{fedd81} 71.49 & \cellcolor[HTML]{fed580} 57.30 & \cellcolor[HTML]{ffea84} 55.65 & \cellcolor[HTML]{fede81} 56.48\\
 & mE5$_{Large}$ & \cellcolor[HTML]{8ac97d} 75.85 & \cellcolor[HTML]{dce182} 45.16 & \cellcolor[HTML]{93cc7e} 60.51 & \cellcolor[HTML]{ffe483} 84.64 & \cellcolor[HTML]{ffe583} 83.11 & \cellcolor[HTML]{ffe483} 83.88 & \cellcolor[HTML]{ffea84} 73.09 & \cellcolor[HTML]{fee182} 71.39 & \cellcolor[HTML]{ffe382} 72.24 & \cellcolor[HTML]{eae583} 60.09 & \cellcolor[HTML]{f1e783} 56.67 & \cellcolor[HTML]{f1e783} 58.38\\
 & E5$_{Large}$ & \cellcolor[HTML]{f2e783} 63.60 & \cellcolor[HTML]{ffe583} 41.52 & \cellcolor[HTML]{fcea84} 52.56 & \cellcolor[HTML]{c3da81} 89.00 & \cellcolor[HTML]{a6d17f} 88.83 & \cellcolor[HTML]{b5d680} 88.92 & \cellcolor[HTML]{d6df82} 75.50 & \cellcolor[HTML]{e3e382} 74.25 & \cellcolor[HTML]{e0e282} 74.88 & \cellcolor[HTML]{fedd81} 58.00 & \cellcolor[HTML]{f6e984} 56.32 & \cellcolor[HTML]{ffe583} 57.16\\
 & Ada-002 & \cellcolor[HTML]{ffe382} 58.97 & \cellcolor[HTML]{ede683} 43.71 & \cellcolor[HTML]{ffe783} 51.34 & \cellcolor[HTML]{b1d47f} 89.71 & \cellcolor[HTML]{9ecf7e} 89.28 & \cellcolor[HTML]{a8d27f} 89.50 & \cellcolor[HTML]{9cce7e} 78.75 & \cellcolor[HTML]{b2d580} 76.86 & \cellcolor[HTML]{a9d27f} 77.81 & \cellcolor[HTML]{f9e984} 59.49 & \cellcolor[HTML]{dce182} 58.03 & \cellcolor[HTML]{eae583} 58.76\\
 & GTE$_{Small}$ & \cellcolor[HTML]{d9e082} 66.62 & \cellcolor[HTML]{f8e984} 42.80 & \cellcolor[HTML]{dfe282} 54.71 & \cellcolor[HTML]{ffe483} 84.62 & \cellcolor[HTML]{ffeb84} 84.22 & \cellcolor[HTML]{ffe683} 84.42 & \cellcolor[HTML]{feda81} 71.19 & \cellcolor[HTML]{fed17f} 69.22 & \cellcolor[HTML]{fed27f} 70.20 & \cellcolor[HTML]{fdbe7b} 55.18 & \cellcolor[HTML]{fdbf7b} 51.86 & \cellcolor[HTML]{fdbd7b} 53.52\\
 & GTE$_{Base}$ & \cellcolor[HTML]{f6e883} 63.21 & \cellcolor[HTML]{ffe984} 41.99 & \cellcolor[HTML]{fbea84} 52.60 & \cellcolor[HTML]{ffeb84} 86.60 & \cellcolor[HTML]{d9e082} 86.22 & \cellcolor[HTML]{ede683} 86.41 & \cellcolor[HTML]{cfdd81} 75.90 & \cellcolor[HTML]{e2e382} 74.30 & \cellcolor[HTML]{dce182} 75.10 & \cellcolor[HTML]{fdcc7e} 56.47 & \cellcolor[HTML]{fede81} 54.62 & \cellcolor[HTML]{fed37f} 55.55\\
 & GTE$_{Large}$ & \cellcolor[HTML]{d6df82} 66.91 & \cellcolor[HTML]{e0e282} 44.79 & \cellcolor[HTML]{d0de81} 55.85 & \cellcolor[HTML]{feeb84} 86.65 & \cellcolor[HTML]{dde182} 86.01 & \cellcolor[HTML]{efe683} 86.33 & \cellcolor[HTML]{c2d980} 76.62 & \cellcolor[HTML]{d5df82} 75.04 & \cellcolor[HTML]{cedd81} 75.83 & \cellcolor[HTML]{feeb84} 59.27 & \cellcolor[HTML]{dfe282} 57.85 & \cellcolor[HTML]{eee683} 58.56\\
 & BGE$_{Small}$ & \cellcolor[HTML]{fedb81} 55.69 & \cellcolor[HTML]{feda81} 39.99 & \cellcolor[HTML]{feda81} 47.84 & \cellcolor[HTML]{ffe984} 86.01 & \cellcolor[HTML]{f0e783} 85.01 & \cellcolor[HTML]{ffeb84} 85.51 & \cellcolor[HTML]{ffea84} 73.04 & \cellcolor[HTML]{fede82} 71.01 & \cellcolor[HTML]{fee182} 72.03 & \cellcolor[HTML]{fed580} 57.31 & \cellcolor[HTML]{feda81} 54.27 & \cellcolor[HTML]{fed680} 55.79\\
 & BGE$_{Base}$ & \cellcolor[HTML]{fed480} 53.14 & \cellcolor[HTML]{e5e483} 44.37 & \cellcolor[HTML]{fedd81} 48.76 & \cellcolor[HTML]{cbdc81} 88.66 & \cellcolor[HTML]{b7d680} 87.98 & \cellcolor[HTML]{c2d980} 88.32 & \cellcolor[HTML]{a3d07f} 78.38 & \cellcolor[HTML]{b3d580} 76.85 & \cellcolor[HTML]{add37f} 77.61 & \cellcolor[HTML]{d5df82} 60.91 & \cellcolor[HTML]{cedd81} 58.92 & \cellcolor[HTML]{d3de81} 59.91\\
 & BGE$_{Large}$ & \cellcolor[HTML]{ffeb84} 62.07 & \cellcolor[HTML]{bdd880} 47.70 & \cellcolor[HTML]{dde182} 54.88 & \cellcolor[HTML]{b4d580} 89.58 & \cellcolor[HTML]{a3d07f} 89.01 & \cellcolor[HTML]{acd37f} 89.30 & \cellcolor[HTML]{8bca7d} 79.70 & \cellcolor[HTML]{94cc7e} 78.46 & \cellcolor[HTML]{91cb7e} 79.08 & \cellcolor[HTML]{9ace7e} \underline{63.29} & \cellcolor[HTML]{90cb7e} \underline{62.88} & \cellcolor[HTML]{95cc7e} \underline{63.09}\\
\midrule
\vl{1em}{12}{\textit{+ Paraphrase and Masking}}  & Instr.$_{Large}$ & \cellcolor[HTML]{fed17f} 52.03 & \cellcolor[HTML]{d0dd81} 46.11 & \cellcolor[HTML]{fede82} 49.07 & \cellcolor[HTML]{a4d17f} 90.22 & \cellcolor[HTML]{9ace7e} 89.49 & \cellcolor[HTML]{a0cf7e} 89.86 & \cellcolor[HTML]{77c47c} \underline{80.82} & \cellcolor[HTML]{81c77d} \underline{79.51} & \cellcolor[HTML]{7cc57c} \underline{80.17} & \cellcolor[HTML]{c6da81} 61.54 & \cellcolor[HTML]{c0d980} 59.79 & \cellcolor[HTML]{c5da81} 60.66\\
 & E5-v2$_{Base}$ & \cellcolor[HTML]{75c37c} \underline{78.39} & \cellcolor[HTML]{f0e783} 43.47 & \cellcolor[HTML]{8eca7d} 60.93 & \cellcolor[HTML]{a1d07f} 90.33 & \cellcolor[HTML]{95cd7e} 89.72 & \cellcolor[HTML]{9cce7e} 90.03 & \cellcolor[HTML]{d0de81} 75.80 & \cellcolor[HTML]{e2e382} 74.31 & \cellcolor[HTML]{dde182} 75.06 & \cellcolor[HTML]{f9e984} 59.48 & \cellcolor[HTML]{dfe282} 57.81 & \cellcolor[HTML]{ece683} 58.65\\
 & E5-v2$_{Large}$ & \cellcolor[HTML]{fed17f} 52.10 & \cellcolor[HTML]{eae583} 44.02 & \cellcolor[HTML]{fedb81} 48.06 & \cellcolor[HTML]{f8e984} 86.88 & \cellcolor[HTML]{ffeb84} 84.24 & \cellcolor[HTML]{ffeb84} 85.56 & \cellcolor[HTML]{dce182} 75.15 & \cellcolor[HTML]{e4e382} 74.22 & \cellcolor[HTML]{e4e382} 74.69 & \cellcolor[HTML]{ece583} 60.02 & \cellcolor[HTML]{d8e082} 58.27 & \cellcolor[HTML]{e2e382} 59.15\\
 & mE5$_{Large}$ & \cellcolor[HTML]{7cc57c} 77.50 & \cellcolor[HTML]{fdca7e} 37.93 & \cellcolor[HTML]{b8d780} 57.72 & \cellcolor[HTML]{ffe683} 85.09 & \cellcolor[HTML]{fede81} 81.62 & \cellcolor[HTML]{fee182} 83.36 & \cellcolor[HTML]{d3de81} 75.68 & \cellcolor[HTML]{e2e382} 74.31 & \cellcolor[HTML]{dee182} 75.00 & \cellcolor[HTML]{d2de81} 61.04 & \cellcolor[HTML]{e1e282} 57.67 & \cellcolor[HTML]{dee282} 59.35\\
 & E5$_{Large}$ & \cellcolor[HTML]{e3e382} 65.37 & \cellcolor[HTML]{ffea84} 42.19 & \cellcolor[HTML]{ece583} 53.78 & \cellcolor[HTML]{78c47c} 91.96 & \cellcolor[HTML]{6bc07b} \underline{91.89} & \cellcolor[HTML]{72c27c} \underline{91.92} & \cellcolor[HTML]{c6da81} 76.40 & \cellcolor[HTML]{c0d980} 76.13 & \cellcolor[HTML]{c6db81} 76.27 & \cellcolor[HTML]{d3de81} 61.01 & \cellcolor[HTML]{cadc81} 59.17 & \cellcolor[HTML]{d0dd81} 60.09\\
 & Ada-002 & \cellcolor[HTML]{cfdd81} 67.81 & \cellcolor[HTML]{cfdd81} 46.22 & \cellcolor[HTML]{c1d980} 57.02 & \cellcolor[HTML]{93cc7e} 90.88 & \cellcolor[HTML]{8dca7d} 90.14 & \cellcolor[HTML]{91cb7e} 90.51 & \cellcolor[HTML]{7dc57c} 80.50 & \cellcolor[HTML]{8bc97d} 78.97 & \cellcolor[HTML]{85c87d} 79.73 & \cellcolor[HTML]{b3d580} 62.30 & \cellcolor[HTML]{bed880} 59.92 & \cellcolor[HTML]{bcd880} 61.11\\
 & GTE$_{Small}$ & \cellcolor[HTML]{cddc81} 68.03 & \cellcolor[HTML]{fed27f} 38.94 & \cellcolor[HTML]{f0e783} 53.48 & \cellcolor[HTML]{d0de81} 88.46 & \cellcolor[HTML]{bbd780} 87.75 & \cellcolor[HTML]{c7db81} 88.11 & \cellcolor[HTML]{fee182} 72.05 & \cellcolor[HTML]{fede81} 70.95 & \cellcolor[HTML]{fedd81} 71.50 & \cellcolor[HTML]{ebe583} 60.04 & \cellcolor[HTML]{e4e382} 57.53 & \cellcolor[HTML]{eae583} 58.78\\
 & GTE$_{Base}$ & \cellcolor[HTML]{63be7b} \textbf{80.50} & \cellcolor[HTML]{bad780} 47.91 & \cellcolor[HTML]{63be7b} \textbf{64.20} & \cellcolor[HTML]{fdea84} 86.68 & \cellcolor[HTML]{efe683} 85.07 & \cellcolor[HTML]{f9e984} 85.88 & \cellcolor[HTML]{cadc81} 76.16 & \cellcolor[HTML]{cfdd81} 75.33 & \cellcolor[HTML]{d0dd81} 75.75 & \cellcolor[HTML]{e9e583} 60.14 & \cellcolor[HTML]{d6df82} 58.41 & \cellcolor[HTML]{e0e282} 59.27\\
 & GTE$_{Large}$ & \cellcolor[HTML]{b1d580} 71.27 & \cellcolor[HTML]{a0d07f} 50.00 & \cellcolor[HTML]{92cb7e} 60.63 & \cellcolor[HTML]{77c47c} \underline{92.00} & \cellcolor[HTML]{75c37c} 91.40 & \cellcolor[HTML]{76c47c} 91.70 & \cellcolor[HTML]{8fcb7e} 79.46 & \cellcolor[HTML]{97cd7e} 78.31 & \cellcolor[HTML]{95cc7e} 78.89 & \cellcolor[HTML]{abd37f} 62.61 & \cellcolor[HTML]{a3d17f} 61.63 & \cellcolor[HTML]{a8d27f} 62.12\\
 & BGE$_{Small}$ & \cellcolor[HTML]{ffeb84} 62.12 & \cellcolor[HTML]{cfdd81} 46.20 & \cellcolor[HTML]{e7e483} 54.16 & \cellcolor[HTML]{8fcb7e} 91.07 & \cellcolor[HTML]{87c87d} 90.45 & \cellcolor[HTML]{8cca7d} 90.76 & \cellcolor[HTML]{d0de81} 75.81 & \cellcolor[HTML]{e3e382} 74.27 & \cellcolor[HTML]{dde182} 75.04 & \cellcolor[HTML]{c6db81} 61.52 & \cellcolor[HTML]{cbdc81} 59.11 & \cellcolor[HTML]{cbdc81} 60.31\\
 & BGE$_{Base}$ & \cellcolor[HTML]{cedd81} 67.91 & \cellcolor[HTML]{a7d27f} 49.46 & \cellcolor[HTML]{abd37f} 58.69 & \cellcolor[HTML]{77c47c} \underline{92.00} & \cellcolor[HTML]{70c27c} 91.63 & \cellcolor[HTML]{74c37c} 91.81 & \cellcolor[HTML]{80c67d} 80.34 & \cellcolor[HTML]{86c87d} 79.25 & \cellcolor[HTML]{83c77d} 79.80 & \cellcolor[HTML]{9fcf7e} 63.09 & \cellcolor[HTML]{9ecf7e} 61.98 & \cellcolor[HTML]{a0d07f} 62.53\\
 & BGE$_{Large}$ & \cellcolor[HTML]{c0d980} 69.57 & \cellcolor[HTML]{82c77d} 52.51 & \cellcolor[HTML]{8cca7d} \underline{61.04} & \cellcolor[HTML]{63be7b} \textbf{92.81} & \cellcolor[HTML]{63be7b} \textbf{92.33} & \cellcolor[HTML]{63be7b} \textbf{92.57} & \cellcolor[HTML]{63be7b} \textbf{81.95} & \cellcolor[HTML]{63be7b} \textbf{81.09} & \cellcolor[HTML]{63be7b} \textbf{81.52} & \cellcolor[HTML]{63be7b} \textbf{65.49} & \cellcolor[HTML]{63be7b} \textbf{65.76} & \cellcolor[HTML]{63be7b} \textbf{65.62}\\
        \bottomrule
    \end{tabular}
    \caption{Performance of baseline and selected models on 4 intent classification tasks. We report accuracy, macro-f1 score and the mean of both for each dataset. For each metric, \textbf{bold} denotes highest score, \underline{underline} denotes second-highest}
    \label{tab:appendix:main-results-expanded}
\end{table*}

\section{Further Ablations}
\label{sec:appendix:further-ablations}

We conduct further ablation studies using \texttt{bge-small-en-v1.5} (Table \ref{tab:appendix:ablations-bge-small}) and \texttt{gte-large} (Table \ref{tab:appendix:ablations-gte-large}) models to verify the findings of our main ablation study conducted on \texttt{bge-large-en-v1.5} (Table \ref{tab:ablations}). We note that similar trends are observed with the different models, in that our proposed setup utilising a combination of the original utterance embedding with paraphrase embedding and masked utterance embedding using entity overlaps produced consistently higher scores.

\begin{table*}[h]
    \centering
    \setlength{\tabcolsep}{2pt}
    \begin{tabular}{l||c|c|c|c||c}
        \toprule
        \textbf{Setup} & \textbf{ATIS} & \textbf{SNIPS} & \textbf{CLINC} & \textbf{MASSIVE} & \textbf{Overall} \\
        \midrule
embeds only & 47.84 & 85.51 & 72.02 & 55.79 & \cellcolor[HTML]{fee182} 65.29\\
pp only & \textbf{55.57} & 84.73 & 71.18 & 59.14 & \cellcolor[HTML]{f8e984} 67.65\\
masked only & 21.77 & 71.66 & 29.94 & 26.66 & \cellcolor[HTML]{f8696b} 37.51\\
embeds + pp & 52.87 & 86.83 & \textbf{75.56} & \underline{60.12} & \cellcolor[HTML]{aed47f} 68.85\\
embeds + masked & 44.11 & 90.53 & 67.12 & 54.01 & \cellcolor[HTML]{fedb81} 63.94\\
pp + masked & 52.44 & \underline{91.16} & 68.17 & 57.95 & \cellcolor[HTML]{ffeb84} 67.43\\
embeds + pp + masked & \underline{54.16} & \textbf{91.19} & 74.47 & 59.82 & \cellcolor[HTML]{6cc17c} \underline{69.91}\\
(overlap) embeds + masked & 44.11 & 90.69 & 69.39 & 55.35 & \cellcolor[HTML]{fee082} 64.89\\
(overlap) pp + masked & 52.44 & 90.68 & 69.41 & 58.32 & \cellcolor[HTML]{f4e883} 67.71\\
(overlap) embeds + pp + masked & \underline{54.16} & 90.76 & \underline{75.04} & \textbf{60.23} & \cellcolor[HTML]{63be7b} \textbf{70.05}\\
        \bottomrule
    \end{tabular}
    \caption{Ablations on 4 intent classification datasets using a \texttt{bge-small-en-v1.5} model. \textbf{Overall} denotes the mean of accuracy and macro-f1 scores across all datasets.}
    \label{tab:appendix:ablations-bge-small}
\end{table*}

\begin{table*}[h]
    \centering
    \setlength{\tabcolsep}{2pt}
    \begin{tabular}{l||c|c|c|c||c}
        \toprule
        \textbf{Setup} & \textbf{ATIS} & \textbf{SNIPS} & \textbf{CLINC} & \textbf{MASSIVE} & \textbf{Overall} \\
        \midrule
embeds only & 55.85 & 86.33 & 75.83 & 58.56 & \cellcolor[HTML]{ffe282} 69.14\\
pp only & 51.39 & 83.93 & 75.87 & 60.49 & \cellcolor[HTML]{fedc81} 67.92\\
masked only & 35.15 & 75.00 & 35.71 & 31.45 & \cellcolor[HTML]{f8696b} 44.33\\
embeds + pp & 55.26 & 86.39 & \underline{78.86} & \textbf{62.29} & \cellcolor[HTML]{ffea84} 70.70\\
embeds + masked & \textbf{61.38} & \textbf{92.34} & 72.92 & 57.10 & \cellcolor[HTML]{ffeb84} 70.94\\
pp + masked & 59.17 & 91.69 & 73.21 & 59.86 & \cellcolor[HTML]{feeb84} 70.98\\
embeds + pp + masked & 60.64 & 91.89 & 78.64 & 61.97 & \cellcolor[HTML]{66bf7b} \underline{73.29}\\
(overlap) embeds + masked & \textbf{61.38} & \underline{92.31} & 74.41 & 57.91 & \cellcolor[HTML]{dce182} 71.50\\
(overlap) pp + masked & 59.17 & 91.42 & 74.33 & 60.06 & \cellcolor[HTML]{ece683} 71.25\\
(overlap) embeds + pp + masked & 60.64 & 91.70 & \textbf{78.89} & \underline{62.14} & \cellcolor[HTML]{63be7b} \textbf{73.34}\\
        \bottomrule
    \end{tabular}
    \caption{Ablations on 4 intent classification datasets using a \texttt{gte-large} model. \textbf{Overall} denotes the mean of accuracy and macro-f1 scores across all datasets.}
    \label{tab:appendix:ablations-gte-large}
\end{table*}

\section{Description Paraphrasing}
\label{sec:appendix:ablations-desc-paraphrases}

To produce paraphrases of intent descriptions, we leverage a \texttt{stablelm-2-1\_6b-chat} model in a similar setup to our inference-time utterance paraphrasal. We increase temperature value from 0.5 to 4.1 in increments of 0.2, producing a paraphrase for each value. We then filter the generated set of descriptions for duplicates and enforce our \textit{Label Preservation} and \textit{Format Consistency} constraints, resulting in an average of 3.94 paraphrases per intent in addition to the original manually produced intent description. Each paraphrase has an average Levenshtein distance of 4.61 to the manual intent description. We replace half of all intent descriptions for each dataset with randomly sampled paraphrases, we produce 200 such combinations and repeat our experiments. Table \ref{tab:appendix:example-paraphrases} shows examples of paraphrased intent deescriptions for each dataset.

\begin{table*}
    \centering
    \small
    \setlength{\tabcolsep}{3pt}
    \begin{tabular}{l|p{4cm}|p{8cm}}
        \toprule
        \textbf{Intent} & \textbf{Description} & \textbf{Paraphrase} \\
        \midrule
\multirow{3}{*}{\texttt{abbreviation}} & \multirow{3}{4cm}{user is asking what an abbreviation stands for or mean} & "user is asking for a definition or explanation of an abbreviation" \\
 & & "user wants clarification on an abbreviation meaning" \\
 & & "user is asking about the meaning of an abbreviation" \\
\midrule
\multirow{3}{*}{\texttt{aircraft}} & \multirow{3}{4cm}{user is asking about an aircraft} & "user is asking about an aircraft ticket or booking details" \\
 & & "user wants to know about an aircraft" \\
 & & "user wants information about an aircraft" \\
\midrule
\multirow{2}{*}{\texttt{airfare}} & \multirow{2}{4cm}{user is asking about fares, costs or airfares} & "user wants to know airfare prices" \\
 & & "user wants to know about airfare prices" \\
\midrule
\multirow{3}{*}{\texttt{AddToPlaylist}} & \multirow{3}{4cm}{user wants to add a song to a playlist} & "user wants to include a song in their playlist" \\
 & & "user wants to incorporate a song into their music collection" \\
 & & "user wants to add a song to their playlist" \\
\midrule
\multirow{3}{*}{\texttt{RateBook}} & \multirow{3}{4cm}{user wants the rating of/to rate a book} & "user wants to give an opinion on a book" \\
 & & "user wants to leave a rating for a book" \\
 & & "user wants to leave a review on/ rate the book" \\
\midrule
\multirow{3}{*}{\texttt{SearchScreeningEvent}} & \multirow{3}{4cm}{user wants to know when a movie is on/screening time of a movie} & "user wants movie screening information" \\
 & & "user wants to know movie screening schedule" \\
 & & "user wants to know movie screening time" \\
\midrule
\multirow{3}{*}{\texttt{accept\_reservations}} & \multirow{3}{4cm}{user wants to know if a location accept reservations} & "user wants to check if the place allows reservations" \\
 & & "user wants to check if a place allows reservations" \\
 & & "user wants to check location reservations" \\
\midrule
\multirow{3}{*}{\texttt{alarm}} & \multirow{3}{4cm}{user wants to set or get an alarm} & "user wants a time alarm" \\
 & & "user wants to set a reminder or schedule an alarm" \\
 & & "user wants to set an alarm clock" \\
\midrule
\multirow{3}{*}{\texttt{calendar}} & \multirow{3}{4cm}{user wants to know about events from their calendar} & "user is asking for event details from their calendar" \\
 & & "user wants to see their calendar for upcoming events" \\
 & & "user wants to check events in their calendar" \\
\midrule
\multirow{3}{*}{\texttt{email\_query}} & \multirow{3}{4cm}{user is asking about email} & "user wants to know how to send an email" \\
 & & "user wants to know how to use email effectively" \\
 & & "user wants an email response or clarification" \\
\midrule
\multirow{3}{*}{\texttt{general\_greet}} & \multirow{3}{4cm}{user is saying a greeting} & "user wants to talk or greet someone" \\
 & & "user wants to greet or say hello" \\
 & & "user wants to greet you or acknowledge your presence" \\
\midrule
\multirow{3}{*}{\texttt{news\_query}} & \multirow{3}{4cm}{user is asking about the news} & "user wants to learn about the latest news" \\
 & & "user wants to know the latest news" \\
 & & "user wants news update or clarification" \\
        \bottomrule
    \end{tabular}
    \caption{Intents, descriptions and example paraphrases from all 4 intent classification datasets.}
    \label{tab:appendix:example-paraphrases}
\end{table*}

\section{t-SNE Visualisation}
\label{sec:appendix:tsne-visualisation}

Due to the challenge to readability posed by the large number of intents in the CLINC dataset, instead sample the 15 top-performing ($100\%$ accuracy) and lowest-performing ($24.47\%$ accuracy) intent classes for illustration, with the results shown in Figures \ref{fig:embedding-visualisation}{c} and \ref{fig:embedding-visualisation}{d} respectively.

\begin{figure*}[ht]
    \centering
    \begin{minipage}{0.48\linewidth}
        \centering
        \includegraphics[width=\linewidth]{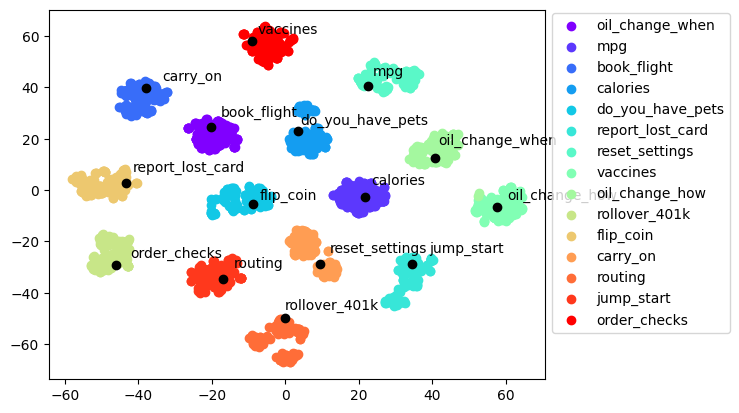}
        \vspace{-25pt}
        \caption*{(\textbf{a})}
    \end{minipage}
    \begin{minipage}{0.48\linewidth}
        \centering
        \includegraphics[width=\linewidth]{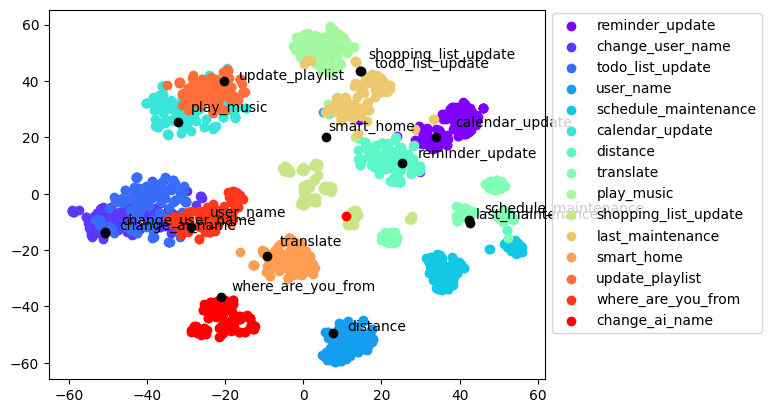}
        \vspace{-25pt}
        \caption*{(\textbf{b})}
    \end{minipage}
    \begin{minipage}{0.48\linewidth}
        \centering
        \includegraphics[width=\linewidth]{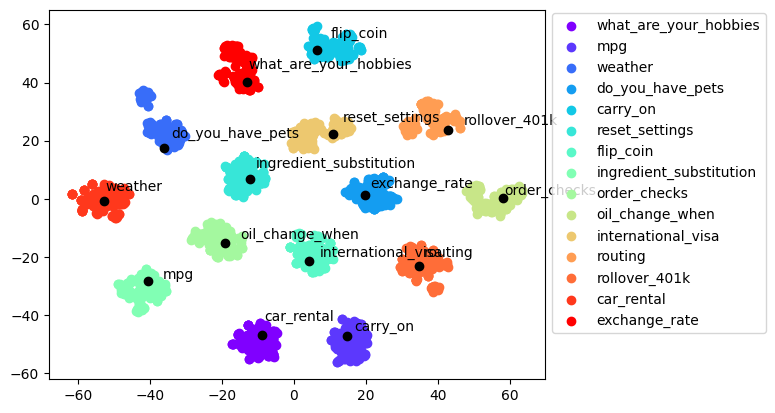}
        \vspace{-25pt}
        \caption*{(\textbf{c})}
    \end{minipage}
    \begin{minipage}{0.48\linewidth}
        \centering
        \includegraphics[width=\linewidth]{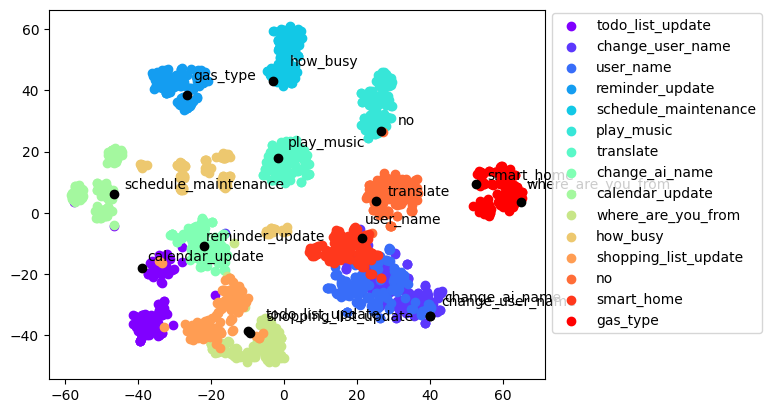}
        \vspace{-25pt}
        \caption*{(\textbf{d})}
    \end{minipage}
    \begin{minipage}{0.48\linewidth}
        \centering
        \includegraphics[width=\linewidth]{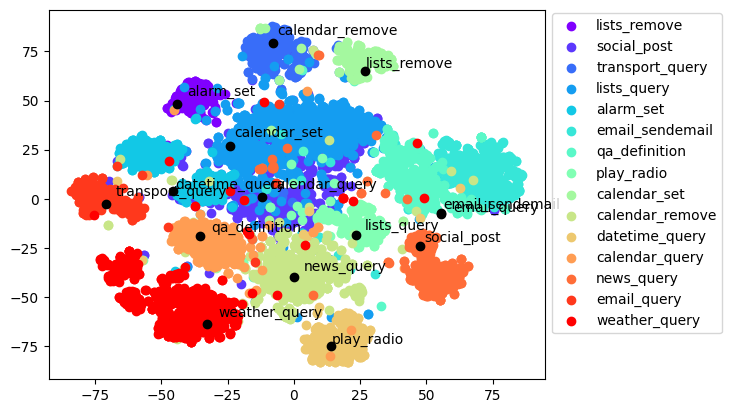}
        \vspace{-25pt}
        \caption*{(\textbf{e})}
    \end{minipage}
    \begin{minipage}{0.48\linewidth}
        \centering
        \includegraphics[width=\linewidth]{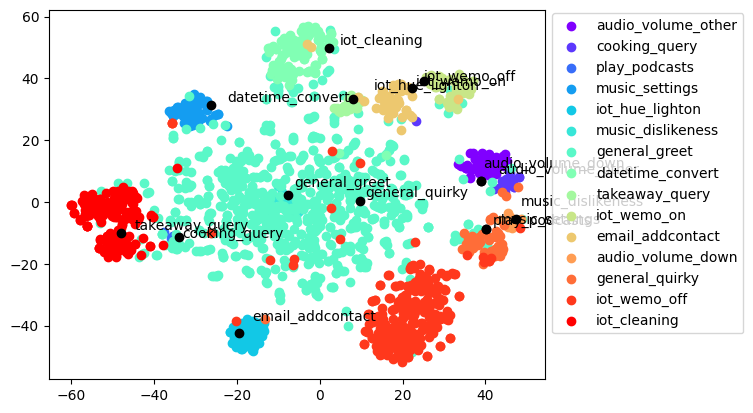}
        \vspace{-25pt}
        \caption*{(\textbf{f})}
    \end{minipage}
    \begin{minipage}{0.48\linewidth}
        \centering
        \includegraphics[width=\linewidth]{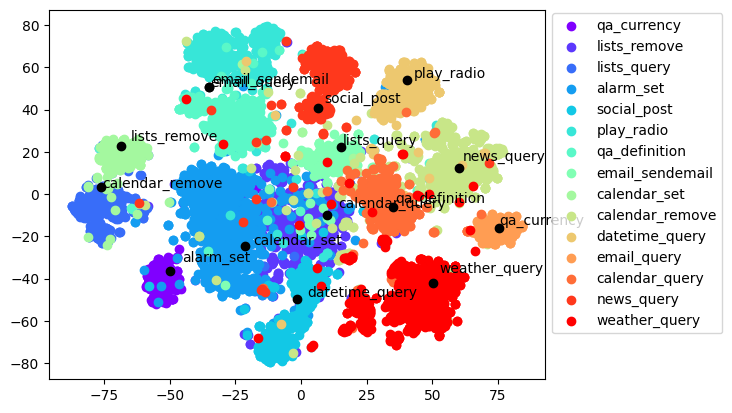}
        \vspace{-25pt}
        \caption*{(\textbf{g})}
    \end{minipage}
    \begin{minipage}{0.48\linewidth}
        \centering
        \includegraphics[width=\linewidth]{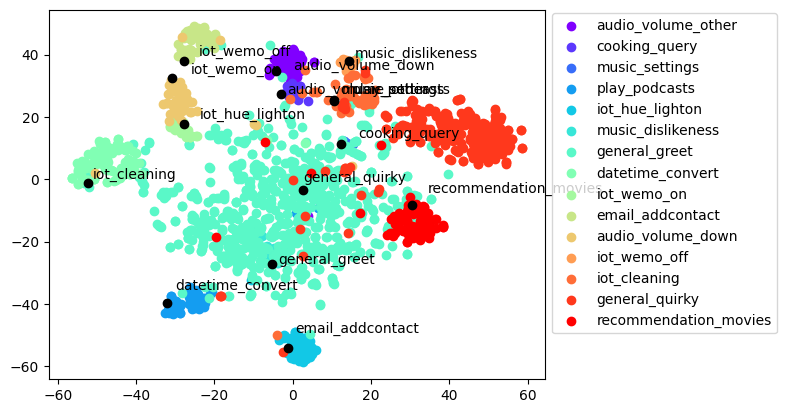}
        \vspace{-25pt}
        \caption*{(\textbf{h})}
    \end{minipage}
    \vspace{-10pt}
    \caption{t-SNE \cite{vandermaaten2008visualizing} visualisation of embeddings for CLINC and MASSIVE datasets computed using BGE$_{Large}$, class label description embeddings are shown in black and labelled. (\textbf{Row 1}) Embeddings of top 15 and bottom 15 classes from CLINC, (\textbf{Row 2}) Embedding + Paraphrasing and Masking of top 15 and bottom 15 classes from CLINC, (\textbf{Row 3}) Embeddings for top 15 and bottom 15 classes from MASSIVE, (\textbf{Row 4}) Embedding + Paraphrasing and Masking of top 15 and bottom 15 classes from CLINC.}
    \label{fig:appendix:tsne-visualisation}
    \vspace{-15pt}
\end{figure*}

\section{Embedding Similarities Analysis}
\label{sec:appendix:embedding-similarity}

\begin{table}[h]
    \centering
    \setlength{\tabcolsep}{3pt}
    \begin{tabular}{l|cc|cc|cc}
        \toprule
         \textbf{Dataset} & $\mu_{s_{in}}$ & $\sigma_{s_{in}}$ & $\mu_{s_{out}} $ & $\sigma_{s_{out}}$ & $\Delta_s$ & $\%\Delta_s$ \\
         \midrule
         ATIS & 0.80 & 0.06 & 0.77 & 0.06 & 0.03 & 3.86 \\
         SNIPS & 0.76 & 0.04 & 0.69 & 0.05 & 0.07 & 8.65 \\
         CLINC & 0.83 & 0.05 & 0.68 & 0.05 & 0.15 & 17.86 \\
         MASSIVE & 0.80 & 0.05 & 0.69 & 0.05 & 0.11 & 13.73 \\
         \bottomrule
    \end{tabular}
    \caption{Mean embedding similarity of sentences within the same class (\textit{in}) and different classes (\textit{out}). $\Delta_s$ denotes the average difference between \textit{in}-class and \textit{out}-class, $\%\Delta_s$ denotes the percentage average difference of similarity.}
    \label{tab:mean-similarity-dataset}
    \vspace{-15pt}
\end{table}

We perform additional analysis on the mean embedding similarity of sentences within the same intent class (\textit{in}-class) and of different intents (\textit{out}-class). For a set of intent classes $\mathcal{C}$ and utterances $\mathcal{U}$, we calculate the mean \textit{in}-class similarity $\mathbf{s}_{in}$  and \textit{out}-class similarity $\mathbf{s}_{out}$ as
\vspace{-10pt}
\begin{equation*}
    \mathbf{s}_{in} = \cfrac{1}{|\mathcal{C}|}\sum_{c\in\mathcal{C}}\sum_{u_i\in\mathcal{U}_c}\sum_{u_j\in\mathcal{U}_c\backslash\{u_i\}}\cfrac{s(\mathbf{h}(u_i), \mathbf{h}(u_{j}))}{n_c(n_c - 1)}
\end{equation*}
\vspace{-5pt}
\begin{equation*}
    \mathbf{s}_{out} = \cfrac{1}{|\mathcal{C}|}\sum_{c\in\mathcal{C}}\sum_{u_i\in\mathcal{U}_c}\sum_{u_j\in\mathcal{U}_{c'}}\cfrac{s(\mathbf{h}(u_i), \mathbf{h}(u_{j}))}{n_cn_{c'}}
\end{equation*}
where $\mathcal{U}_c$ and $\mathcal{U}_{c'}$ denotes the set of utterances belonging to class $c$ and all classes other than $c'$ respectively, $n_c$ is the number of utterances in set $\mathcal{U}_c$. The mean \textit{in}-class and \textit{out}-class similarity scores are shown per dataset (Table \ref{tab:mean-similarity-dataset}). From a basic correlation analysis of the mean embedding similarity against a number of metrics, we note for model performance on the MTEB benchmark there exists a strong positive correlation to the difference $\Delta_s$ between \textit{in}-class and \textit{out}-class examples (Pearson $r=0.72$, $p<0.01$) as well as $\%\Delta_s$ (Pearson $r=0.73$, $p<0.01$), and there exists a strong negative correlation to the mean \textit{out}-class similarity $\mu_{s_{out}}$ (Pearson $r=-0.72$, $p<0.01$).

\section{Synthetic Examples}
\label{sec:appendix:intents-desc-examples}

We compare additionally against synthetic utterance generated for each intent class. We leverage \texttt{gpt-3.5-turbo} \cite{openai2023gpt4} for this purpose, by including the tokenized intent labels and label description within the prompt to generate a set $\mathcal{S}$ of questions or commands fitting said intent i.e. ``\textit{Given a category} \texttt{tokenized\_intent} \textit{and the description} \texttt{description}\textit{, Please generate} \texttt{n} \textit{different example sentences of users asking questions or making commands that fit the given category.}". At inference time, we sample $k$ synthetic examples for $c$ classes and make prediction $\hat{y}_i$ as follows:

\vspace{-20pt}
\begin{equation*}
    \hat{y}_i = \underset{c}{\arg\max}\cfrac{\sum^k_ms(\mathbf{h}(u_i), \mathbf{h}(s^c_m))}{k}
\end{equation*}
\vspace{-10pt}

\noindent where $s^c_m$ denotes the $m^{th}$ example utterance belonging to intent class $c\in\mathcal{C}$. Examples of synthetic utterances can be found in Appendix \ref{sec:appendix:intents-desc-examples}. We report on the results separately in Section \ref{sec:results:synthetic} and the full results can be seen in Appendix \ref{sec:appendix:synthetic-full}. We also consider synthetic examples generated using \texttt{gpt-4} but found the average performance to be lower on our task (Appendix \ref{sec:appendix:synthetic-gpt4}).

\subsection{Results: Methods using Synthetic Data}
\label{sec:results:synthetic}

\begin{table}
    \centering
    \small
    \setlength{\tabcolsep}{3pt}
    \begin{tabular}{c| >{\normalsize} l|c|c|c|c|c|c}
    \toprule
    \multirow{2}{*}{$k$}& \multirow{2}{*}{\normalsize \textbf{Metric}} & \multicolumn{2}{c |}{\normalsize\textbf{ATIS}} & \multicolumn{2}{c |}{\normalsize\textbf{SNIPS}} & \multicolumn{2}{c }{\normalsize\textbf{CLINC}} \\
    \cline{3-8}
    & & \normalsize $\mu$ & \normalsize $\sigma$ & \normalsize $\mu$ & \normalsize $\sigma$ & \normalsize $\mu$ & \normalsize $\sigma$ \\
    \midrule
\vl{1em}{3}{\textit{$k=1$}}  & Mean & 23.59 & 8.42 & 71.37 & 5.51 & 53.87 & 5.42\\
 & $\Delta_{Label}$ & -6.15 & -4.23 & -4.94 & -1.02 & -13.31 & 0.37\\
 & $\Delta_{Desc}$ & -24.08 & 4.38 & -15.54 & 2.57 & -20.60 & 2.48\\
\midrule
\vl{1em}{3}{\textit{$k=3$}}  & Mean & 28.63 & 7.41 & 77.27 & 4.16 & 64.65 & 3.21\\
 & $\Delta_{Label}$ & -1.10 & -5.23 & 0.96 & -2.37 & -2.53 & -1.84\\
 & $\Delta_{Desc}$ & -19.03 & 3.37 & -9.64 & 1.22 & -9.82 & 0.27\\
\midrule
\vl{1em}{3}{\textit{$k=5$}}  & Mean & 30.05 & 6.74 & 78.54 & 3.98 & 67.29 & 2.81\\
 & $\Delta_{Label}$ & 0.31 & -5.90 & 2.24 & -2.55 & 0.11 & -2.23\\
 & $\Delta_{Desc}$ & -17.62 & 2.70 & -8.36 & 1.04 & -7.18 & -0.13\\
\midrule
\vl{1em}{3}{\textit{$k=10$}}  & Mean & 30.80 & 5.33 & 79.63 & 3.57 & 69.24 & 2.48\\
 & $\Delta_{Label}$ & 1.06 & -7.31 & 3.32 & -2.96 & 2.06 & -2.57\\
 & $\Delta_{Desc}$ & -16.87 & 1.29 & -7.28 & 0.63 & -5.23 & -0.46\\
\midrule
\vl{1em}{3}{\textit{$k=15$}}  & Mean & 31.12 & 5.15 & 80.06 & 3.46 & 69.99 & 2.50\\
 & $\Delta_{Label}$ & 1.38 & -7.49 & 3.75 & -3.07 & 2.80 & -2.55\\
 & $\Delta_{Desc}$ & -16.55 & 1.12 & -6.85 & 0.52 & -4.49 & -0.44\\
    \bottomrule
    \end{tabular}
    \caption{Averaged mean of accuracy and macro-f1 scores experiments conducted across 20 samples and 12 models using $k$ number of synthetic examples per intent class. $\Delta_{Label}$ and $\Delta_{Desc}$ are differences to the averaged performance of methods using tokenized labels and intent descriptions respectively.}
    \label{tab:synthetic-examples}
    \vspace{-15pt}
\end{table}

We evaluate the efficacy of methods using synthetic examples by generating a set of $n=20$ synthetic examples, from which we sample $k$ to act as class prototypes, we repeat this procedure 20 times and compute the average performance across all samples. Table \ref{tab:synthetic-examples} shows averaged model performance across all 12 selected models and samples for $k = [1, 3, 5, 10, 15]$. For full results see Table \ref{tab:appendix:synthetic-full} in Appendix \ref{sec:appendix:synthetic-full}. We conducted additional experimentation with $k > 15$ but found further increasing $k$ did not yield significant improvements in performance. We note our method using $k = 15$ synthetic examples outperforms tokenized labels on SNIPS (80.06 vs 76.30) and CLINC (69.99 vs 67.18) datasets, but underperforms slightly on the ATIS dataset (31.12 vs 31.70). Synthetic examples underperforms description-based methods by a considerable margin on all datasets, suggesting single intent label descriptions can be more powerful as class prototypes than synthetic instances. We note also the higher standard deviation $\sigma$ in performance compared to the description-augmented method but lower compared to methods using tokenized labels.

\subsection{Table of intents, descriptions and sampled synthetic examples generated using gpt-3.5-turbo}
See Table \ref{tab:appendix:intents-atis} (ATIS), Table \ref{tab:appendix:intents-snips} (SNIPS) and Table \ref{tab:appendix:intents-clinc} (CLINC).

\begin{table*}
    \centering
    \small
    \setlength{\tabcolsep}{2pt}
    \begin{tabular}{l|p{3cm}|p{12cm}}
        \toprule
        \textbf{Intent} & \textbf{Description} & \textbf{Synthetic Examples} \\
        \midrule
        \multirow{3}{*}{abbreviation} & \multirow{3}{3cm}{user is asking what an abbreviation stands for or mean} & ``what does eta stand for?" \\
 &  & ``can you tell me the meaning of atc?" \\
 &  & ``what is the abbreviation vfr referring to?" \\
\midrule
\multirow{3}{*}{aircraft} & \multirow{3}{3cm}{user is asking about an aircraft} & ``what is the maximum speed of this aircraft?" \\
 &  & ``can you provide me with the dimensions of the aircraft?" \\
 &  & ``how many passengers can this aircraft accommodate?" \\
\midrule
\multirow{3}{*}{airfare} & \multirow{3}{3cm}{user is asking about fares, costs or airfares} & ``what are the airfare options for a round-trip flight from new york to los angeles?" \\
 &  & ``can you provide me with the cost of a first-class airfare from london to paris?" \\
 &  & ``how much does it usually cost for a one-way airfare from tokyo to sydney?" \\
\midrule
\multirow{3}{*}{airline} & \multirow{3}{3cm}{user is asking about an airline/airlines} & ``which airline offers the most affordable tickets from los angeles to new york?" \\
 &  & ``can you recommend any airlines that provide extra legroom for tall passengers?" \\
 &  & ``what are the baggage restrictions for this airline?" \\
\midrule
\multirow{3}{*}{airport} & \multirow{3}{3cm}{user is asking about an airport/airports} & ``which airports in new york have direct flights to los angeles?" \\
 &  & ``can you provide me with information about the nearest airport to my current location?" \\
 &  & ``how long does it take to get from the city center to heathrow airport?" \\
\midrule
\multirow{3}{*}{capacity} & \multirow{3}{3cm}{user is asking about capacity (of an aircraft)} & ``what is the seating capacity of a boeing 747 aircraft?" \\
 &  & ``can you tell me the maximum passenger capacity of a airbus a380?" \\
 &  & ``what is the cargo capacity of a cessna 172 aircraft?" \\
\midrule
\multirow{3}{*}{cheapest} & \multirow{3}{3cm}{user is asking about the cheapest (fare)} & ``can you find me the cheapest flight from new york to los angeles?" \\
 &  & ``i need the cheapest airfare available for a one-way trip from london to barcelona." \\
 &  & ``what is the cheapest flight i can get from chicago to miami during the christmas holidays?" \\
\midrule
\multirow{3}{*}{city} & \multirow{3}{3cm}{user is asking about a city or place} & ``can you provide me with flight options to new york city?" \\
 &  & ``what are the popular attractions in san francisco?" \\
 &  & ``which airlines operate flights to tokyo?" \\
\midrule
\multirow{3}{*}{day\_name} & \multirow{3}{3cm}{user is asking about a day (of the week)} & ``which day of the week is the best to book a flight?" \\
 &  & ``can you tell me the day of the week for my flight to new york?" \\
 &  & ``what is the departure day for the flight to london?" \\
\midrule
\multirow{3}{*}{distance} & \multirow{3}{3cm}{user is asking for the distance between places/locations} & ``what is the distance between new york and los angeles?" \\
 &  & ``calculate the distance from london to paris." \\
 &  & ``how far is it from sydney to melbourne?" \\
\midrule
\multirow{3}{*}{flight} & \multirow{3}{3cm}{user is asking about available flights} & ``what flights are available from new york city to los angeles tomorrow?" \\
 &  & ``can you please check if there are any direct flights from london to tokyo?" \\
 &  & ``i need to book a one-way flight from chicago to miami on the 15th of june." \\
\midrule
\multirow{3}{*}{flight\_no} & \multirow{3}{3cm}{user is asking about a flight number} & ``what is the flight number for the flight from new york to london?" \\
 &  & ``can you provide me with the flight number for the 6:00 am departure to los angeles?" \\
 &  & ``i need to know the flight number for the red-eye flight to chicago." \\
\midrule
\multirow{3}{*}{flight\_time} & \multirow{3}{3cm}{user is asking about departue time or schedule for a flight} & ``what is the flight time for the next available flight to new york?" \\
 &  & ``can you tell me the departure time for flight 123 to london?" \\
 &  & ``i need to know the schedule for flights leaving tomorrow morning." \\
\midrule
\multirow{3}{*}{ground\_fare} & \multirow{3}{3cm}{user is asking about the ground fare at a destination} & ``what is the average ground fare in los angeles?" \\
 &  & ``can you provide information about ground fares in paris?" \\
 &  & ``how much should i expect to pay for ground transportation in london?" \\
\midrule
\multirow{3}{*}{ground\_service} & \multirow{3}{3cm}{user is asking about ground service at a location} & ``what are the available ground services at this airport?" \\
 &  & ``can you provide me with information about ground services at the destination airport?" \\
 &  & ``is there wheelchair assistance available as part of the ground services?" \\
\midrule
\multirow{3}{*}{meal} & \multirow{3}{3cm}{user is asking about meals/catering} & ``what meal options are available for the flight?" \\
 &  & ``can i request a vegetarian meal for my flight?" \\
 &  & ``do you have any special meals for passengers with dietary restrictions?" \\
\midrule
\multirow{3}{*}{quantity} & \multirow{3}{3cm}{user is asking about the quantity/amount of something} & ``how many flight attendants are there on this flight?" \\
 &  & ``could you tell me the total weight of the luggage allowed per passenger?" \\
 &  & ``how many passengers are currently on board the plane?" \\
\midrule
\multirow{3}{*}{restriction} & \multirow{3}{3cm}{user is asking about restrictions} & ``can you please provide me with the baggage restrictions for my upcoming flight?" \\
 &  & ``what are the restrictions on carrying liquids in my hand luggage?" \\
 &  & ``are there any age restrictions for children traveling alone on your flights?" \\
        \bottomrule
    \end{tabular}
    \caption{Intents, descriptions and synthetic examples for the ATIS dataset.}
    \label{tab:appendix:intents-atis}
\end{table*}

\begin{table*}
    \centering
    \small
    \setlength{\tabcolsep}{2pt}
    \begin{tabular}{l|p{4cm}|p{8cm}}
        \toprule
        \textbf{Intent} & \textbf{Description} & \textbf{Synthetic Examples} \\
        \midrule
        \multirow{3}{*}{AddToPlaylist} & \multirow{3}{4cm}{user wants to add a song to a playlist} & ``hey, can you please add this new release to my workout playlist?" \\
 &  & ``add the latest hit by taylor swift to my party playlist, please." \\
 &  & ``can you include this classic rock track in my road trip playlist?" \\
\midrule
\multirow{3}{*}{BookRestaurant} & \multirow{3}{4cm}{user wants to book/make a reservation at a restaurant} & ``can you help me book a table at a fancy restaurant for this saturday?" \\
 &  & ``i would like to make a reservation for two at the most popular restaurant in town." \\
 &  & ``what is the best way to book a restaurant online?" \\
\midrule
\multirow{3}{*}{GetWeather} & \multirow{3}{4cm}{user wants to know about the weather} & ``what will be the weather like tomorrow?" \\
 &  & ``can you provide me with a detailed weather forecast for the next week?" \\
 &  & ``is it going to rain today?" \\
\midrule
\multirow{3}{*}{PlayMusic} & \multirow{3}{4cm}{user wants to play a song} & ``hey, playmusic! can you play 'shape of you' by ed sheeran?" \\
 &  & ``playmusic, please play some soothing music to help me relax." \\
 &  & ``i'm in the mood for some throwback tunes. playmusic, can you play 'don't stop believin' by journey?" \\
\midrule
\multirow{3}{*}{RateBook} & \multirow{3}{4cm}{user wants the rating of/to rate a book} & ``can anyone recommend a ratebook website where I can find reviews and ratings for the latest bestsellers?" \\
 &  & ``what's the highest-rated ratebook on the market right now? i want to make sure i'm picking something worthwhile." \\
 &  & ``i'd like some suggestions for popular ratebooks in the fantasy genre. any recommendations?" \\
\midrule
\multirow{3}{*}{SearchCreativeWork} & \multirow{3}{4cm}{user wants to find a creative work (book, song etc.)} & ``can you help me search for a creative work that is similar to harry potter?" \\
 &  & ``i'm looking for a book recommendation, search for a creative work with a thrilling mystery plot." \\
 &  & ``find me a song that has won multiple awards and has a catchy melody." \\
\midrule
\multirow{3}{*}{SearchScreeningEvent} & \multirow{3}{4cm}{user wants to know when a movie is on/screening time of a movie} & ``when is the next screening event for the movie avengers: endgame?" \\
 &  & ``what are the screening times for the romantic comedy crazy, stupid, love?" \\
 &  & ``can you tell me the showtimes for the movie joker in theaters nearby?" \\
        \bottomrule
    \end{tabular}
    \caption{Intents, descriptions and synthetic examples for the SNIPS dataset.}
    \label{tab:appendix:intents-snips}
\end{table*}

\begin{table*}
    \centering
    \small
    \setlength{\tabcolsep}{2pt}
    \begin{tabular}{l|p{4cm}|p{8cm}}
        \toprule
        \textbf{Intent} & \textbf{Description} & \textbf{Synthetic Examples} \\
        \midrule
\multirow{3}{*}{timezone} & \multirow{3}{3cm}{user is asking about timezone} & ``can you please tell me the current timezone in new york city?" \\
 &  & ``what is the timezone difference between san francisco and tokyo?" \\
 &  & ``i need to know the exact timezone utc offset for london." \\
\midrule
\multirow{3}{*}{fun\_fact} & \multirow{3}{3cm}{user wants to know a fun fact} & ``tell me a fun fact about the eiffel tower!" \\
 &  & ``i need a fun fact to impress my friends at dinner tonight." \\
 &  & ``give me a fun fact about dolphins, please." \\
\midrule
\multirow{3}{*}{date} & \multirow{3}{3cm}{user wants to know the current date} & ``what is today's date?" \\
 &  & ``can you tell me the current date?" \\
 &  & ``i need to know the date, please." \\
\midrule
\multirow{3}{*}{what\_can\_i\_ask\_you} & \multirow{3}{3cm}{user wants to know what can be asked} & ``what kind of questions are appropriate to ask you?" \\
 &  & ``can you provide me with a list of topics that i can ask you about?" \\
 &  & ``i'm curious, what type of questions am i allowed to ask in this category?" \\
\midrule
\multirow{3}{*}{todo\_list\_update} & \multirow{3}{3cm}{user wants to update or change todo list} & ``how can i edit my current todo list?" \\
 &  & ``can you show me how to modify my existing tasks in the todo list?" \\
 &  & ``is there a way to change the priority of items on the todo list?" \\
\midrule
\multirow{3}{*}{bill\_balance} & \multirow{3}{3cm}{user wants to know their bill balance} & ``what is my current bill balance?" \\
 &  & ``can you please provide the details of my bill balance?" \\
 &  & ``i need to know how much is due on my bill." \\
\midrule
\multirow{3}{*}{schedule\_meeting} & \multirow{3}{3cm}{user wants to schedule meeting} & ``can you help me schedule a meeting for next week?" \\
 &  & ``i need assistance in setting up a meeting with our new client." \\
 &  & ``how do i go about scheduling a team meeting for tomorrow?" \\
\midrule
\multirow{3}{*}{routing} & \multirow{3}{3cm}{user wants to know about routing number} & ``what is a routing number and why is it important for banking?" \\
 &  & ``how can i find the routing number for my bank account?" \\
 &  & ``can you explain the specific purpose of a routing number in online transactions?" \\
\midrule
\multirow{3}{*}{food\_last} & \multirow{3}{3cm}{user wants to know how long a food lasts} & ``how long can i safely keep cooked chicken in the refrigerator?" \\
 &  & ``what is the shelf life of fresh milk at room temperature?" \\
 &  & ``can you give me some tips on how to extend the life of avocados?" \\
\midrule
\multirow{3}{*}{bill\_due} & \multirow{3}{3cm}{user wants to know when a bill is due} & ``hey, can you remind me when my electricity bill is due?" \\
 &  & ``what's the due date for my credit card bill this month?" \\
 &  & ``i need to know when my phone bill is due. can you help me with that?" \\
\midrule
\multirow{3}{*}{time} & \multirow{3}{3cm}{user is asking for the time} & ``what is the current time?" \\
 &  & ``could you please tell me what time it is?" \\
 &  & ``do you have the time?" \\
\midrule
\multirow{3}{*}{freeze\_account} & \multirow{3}{3cm}{user wants to freeze their account} & ``how can i freeze my account temporarily?" \\
 &  & ``i need to put a hold on my account, can you assist me?" \\
 &  & ``please freeze my account until further notice." \\
\midrule
\multirow{3}{*}{rollover\_401k} & \multirow{3}{3cm}{user wants to know about 401k rollover} & ``how can i rollover my 401k into a new retirement account?" \\
 &  & ``can you explain the process of a 401k rollover to me?" \\
 &  & ``what are the benefits of doing a rollover with my 401k?" \\
\midrule
\multirow{3}{*}{travel\_alert} & \multirow{3}{3cm}{user wants to know about travel alerts} & ``are there any current travel alerts that i should be aware of?" \\
 &  & ``notify me if there are any travel alerts for my upcoming destination." \\
 &  & ``can you provide me with the latest travel alerts for international travel?" \\
\midrule
\multirow{3}{*}{translate} & \multirow{3}{3cm}{user wants to translate} & ``can you translate this document from english to french?" \\
 &  & ``excuse me, i need assistance translating this menu into spanish." \\
 &  & ``how can i translate this phrase into italian?" \\
        \bottomrule
    \end{tabular}
    \caption{Intents, descriptions and synthetic examples for 15 intents from the CLINC dataset.}
    \label{tab:appendix:intents-clinc}
\end{table*}

\section{Full table of results for approach using synthetic examples generated using gpt-3.5-turbo}
\label{sec:appendix:synthetic-full}
See Table \ref{tab:appendix:synthetic-full}.

\begin{table*}
    \centering
    \small
	\setlength{\tabcolsep}{2pt}
    \begin{tabular}{c|l||c|c|c||c|c|c||c|c|c}
        \toprule
        & \multirow{2}{*}{\textbf{Model}} & \multicolumn{3}{c||}{\textbf{ATIS}} & \multicolumn{3}{c||}{\textbf{SNIPS}} & \multicolumn{3}{c}{\textbf{CLINC}} \\
        \cline{3-11}
        & & Acc & F1 & Mean & Acc & F1 & Mean & Acc & F1 & Mean \\
        \midrule
\vl{1em}{12}{\textit{$n=1$}}  & InstructOR$_{Large}$ & \cellcolor[HTML]{fdce7e} 32.77 & \cellcolor[HTML]{fba877} 23.99 & \cellcolor[HTML]{fcb87a} 28.38 & \cellcolor[HTML]{fdc47d} 72.60 & \cellcolor[HTML]{fed07f} 69.26 & \cellcolor[HTML]{fdca7e} 70.93 & \cellcolor[HTML]{fcaa78} 56.94 & \cellcolor[HTML]{fba677} 53.71 & \cellcolor[HTML]{fba877} 55.32\\
 & E5-v2$_{Base}$ & \cellcolor[HTML]{fa8d72} 27.01 & \cellcolor[HTML]{f9736d} 19.30 & \cellcolor[HTML]{f97f6f} 23.16 & \cellcolor[HTML]{fcac78} 70.28 & \cellcolor[HTML]{fcb079} 66.52 & \cellcolor[HTML]{fcae78} 68.40 & \cellcolor[HTML]{f98270} 50.05 & \cellcolor[HTML]{f97f6f} 47.21 & \cellcolor[HTML]{f98170} 48.63\\
 & E5-v2$_{Large}$ & \cellcolor[HTML]{fba977} 29.50 & \cellcolor[HTML]{f8716d} 19.12 & \cellcolor[HTML]{fa8c72} 24.31 & \cellcolor[HTML]{fa9473} 68.09 & \cellcolor[HTML]{fb9874} 64.41 & \cellcolor[HTML]{fa9674} 66.25 & \cellcolor[HTML]{f8726d} 47.24 & \cellcolor[HTML]{f86f6c} 44.54 & \cellcolor[HTML]{f8716c} 45.89\\
 & Multilingual-E5$_{Large}$ & \cellcolor[HTML]{f8696b} 23.85 & \cellcolor[HTML]{f8696b} 18.37 & \cellcolor[HTML]{f8696b} 21.11 & \cellcolor[HTML]{f8696b} 64.02 & \cellcolor[HTML]{f8696b} 60.24 & \cellcolor[HTML]{f8696b} 62.13 & \cellcolor[HTML]{f8696b} 45.68 & \cellcolor[HTML]{f8696b} 43.54 & \cellcolor[HTML]{f8696b} 44.61\\
 & E5$_{Large}$ & \cellcolor[HTML]{fb9e75} 28.57 & \cellcolor[HTML]{f97e6f} 20.22 & \cellcolor[HTML]{fa8d72} 24.40 & \cellcolor[HTML]{fba276} 69.35 & \cellcolor[HTML]{fcac78} 66.13 & \cellcolor[HTML]{fba777} 67.74 & \cellcolor[HTML]{fb9c75} 54.44 & \cellcolor[HTML]{fb9874} 51.38 & \cellcolor[HTML]{fb9a74} 52.91\\
 & OpenAI-Ada-002 & \cellcolor[HTML]{fcb87a} 30.86 & \cellcolor[HTML]{f9746d} 19.40 & \cellcolor[HTML]{fa9573} 25.13 & \cellcolor[HTML]{fee182} 75.35 & \cellcolor[HTML]{f5e883} 72.78 & \cellcolor[HTML]{feeb84} 74.07 & \cellcolor[HTML]{fcaf78} 57.70 & \cellcolor[HTML]{fbaa77} 54.42 & \cellcolor[HTML]{fcad78} 56.06\\
 & GTE$_{Small}$ & \cellcolor[HTML]{f9806f} 25.87 & \cellcolor[HTML]{f97d6f} 20.15 & \cellcolor[HTML]{f97e6f} 23.01 & \cellcolor[HTML]{f9786e} 65.42 & \cellcolor[HTML]{f97f6f} 62.17 & \cellcolor[HTML]{f97b6f} 63.80 & \cellcolor[HTML]{fa8a71} 51.37 & \cellcolor[HTML]{fa8671} 48.41 & \cellcolor[HTML]{fa8871} 49.89\\
 & GTE$_{Base}$ & \cellcolor[HTML]{f97a6e} 25.34 & \cellcolor[HTML]{f97f6f} 20.33 & \cellcolor[HTML]{f97c6f} 22.83 & \cellcolor[HTML]{fb9f75} 69.09 & \cellcolor[HTML]{fba977} 65.89 & \cellcolor[HTML]{fba476} 67.49 & \cellcolor[HTML]{fa9473} 53.10 & \cellcolor[HTML]{fa9072} 50.04 & \cellcolor[HTML]{fa9273} 51.57\\
 & GTE$_{Large}$ & \cellcolor[HTML]{fcae78} 29.94 & \cellcolor[HTML]{fa9072} 21.83 & \cellcolor[HTML]{fb9d75} 25.88 & \cellcolor[HTML]{fba977} 70.02 & \cellcolor[HTML]{fcb179} 66.56 & \cellcolor[HTML]{fcad78} 68.29 & \cellcolor[HTML]{fb9f75} 54.95 & \cellcolor[HTML]{fb9a74} 51.72 & \cellcolor[HTML]{fb9d75} 53.34\\
 & BGE$_{Small}$ & \cellcolor[HTML]{fa9273} 27.44 & \cellcolor[HTML]{fa8a71} 21.32 & \cellcolor[HTML]{fa8c72} 24.38 & \cellcolor[HTML]{f98470} 66.60 & \cellcolor[HTML]{fa8671} 62.76 & \cellcolor[HTML]{fa8570} 64.68 & \cellcolor[HTML]{fa9273} 52.69 & \cellcolor[HTML]{fa8d72} 49.56 & \cellcolor[HTML]{fa8f72} 51.13\\
 & BGE$_{Base}$ & \cellcolor[HTML]{f8716d} 24.57 & \cellcolor[HTML]{f98270} 20.62 & \cellcolor[HTML]{f9796e} 22.59 & \cellcolor[HTML]{fcad78} 70.39 & \cellcolor[HTML]{fcb079} 66.52 & \cellcolor[HTML]{fcae78} 68.46 & \cellcolor[HTML]{fba176} 55.24 & \cellcolor[HTML]{fb9d75} 52.21 & \cellcolor[HTML]{fb9f75} 53.72\\
 & BGE$_{Large}$ & \cellcolor[HTML]{fedc81} 33.97 & \cellcolor[HTML]{fba677} 23.83 & \cellcolor[HTML]{fdbd7b} 28.90 & \cellcolor[HTML]{fcb67a} 71.31 & \cellcolor[HTML]{fcb97a} 67.29 & \cellcolor[HTML]{fcb87a} 69.30 & \cellcolor[HTML]{fcb279} 58.17 & \cellcolor[HTML]{fcac78} 54.73 & \cellcolor[HTML]{fcaf78} 56.45\\
\midrule
\vl{1em}{12}{\textit{$n=3$}}  & InstructOR$_{Large}$ & \cellcolor[HTML]{e6e483} 39.20 & \cellcolor[HTML]{ffe282} 29.25 & \cellcolor[HTML]{f7e984} 34.22 & \cellcolor[HTML]{fae984} 76.71 & \cellcolor[HTML]{f9e984} 72.39 & \cellcolor[HTML]{f9e984} 74.55 & \cellcolor[HTML]{ffea84} 67.88 & \cellcolor[HTML]{ffe883} 64.84 & \cellcolor[HTML]{ffe984} 66.36\\
 & E5-v2$_{Base}$ & \cellcolor[HTML]{fcea84} 35.75 & \cellcolor[HTML]{fdc97d} 26.97 & \cellcolor[HTML]{fed880} 31.36 & \cellcolor[HTML]{ffeb84} 76.25 & \cellcolor[HTML]{ffea84} 71.56 & \cellcolor[HTML]{ffea84} 73.90 & \cellcolor[HTML]{fed17f} 63.52 & \cellcolor[HTML]{fdcf7f} 60.63 & \cellcolor[HTML]{fed07f} 62.08\\
 & E5-v2$_{Large}$ & \cellcolor[HTML]{dee182} 40.41 & \cellcolor[HTML]{fed37f} 27.85 & \cellcolor[HTML]{f7e984} 34.13 & \cellcolor[HTML]{ffe583} 75.68 & \cellcolor[HTML]{ffe382} 70.98 & \cellcolor[HTML]{ffe483} 73.33 & \cellcolor[HTML]{fdca7e} 62.35 & \cellcolor[HTML]{fdc87d} 59.47 & \cellcolor[HTML]{fdc97e} 60.91\\
 & Multilingual-E5$_{Large}$ & \cellcolor[HTML]{f9776e} 25.07 & \cellcolor[HTML]{fdbd7b} 25.90 & \cellcolor[HTML]{fb9874} 25.48 & \cellcolor[HTML]{ffe583} 75.67 & \cellcolor[HTML]{ffe382} 70.93 & \cellcolor[HTML]{ffe483} 73.30 & \cellcolor[HTML]{fdbf7c} 60.56 & \cellcolor[HTML]{fdc07c} 58.19 & \cellcolor[HTML]{fdc07c} 59.37\\
 & E5$_{Large}$ & \cellcolor[HTML]{f2e783} 37.33 & \cellcolor[HTML]{ffe683} 29.64 & \cellcolor[HTML]{fcea84} 33.48 & \cellcolor[HTML]{fed981} 74.57 & \cellcolor[HTML]{fedb81} 70.24 & \cellcolor[HTML]{feda81} 72.40 & \cellcolor[HTML]{ffe683} 67.18 & \cellcolor[HTML]{ffe483} 64.25 & \cellcolor[HTML]{ffe683} 65.72\\
 & OpenAI-Ada-002 & \cellcolor[HTML]{b3d580} 46.96 & \cellcolor[HTML]{fdc47c} 26.53 & \cellcolor[HTML]{e3e382} 36.74 & \cellcolor[HTML]{b7d680} 82.42 & \cellcolor[HTML]{b2d580} 80.27 & \cellcolor[HTML]{b4d580} 81.34 & \cellcolor[HTML]{f5e883} 68.77 & \cellcolor[HTML]{faea84} 65.77 & \cellcolor[HTML]{f7e984} 67.27\\
 & GTE$_{Small}$ & \cellcolor[HTML]{f8706c} 24.50 & \cellcolor[HTML]{fdc87d} 26.95 & \cellcolor[HTML]{fb9b75} 25.72 & \cellcolor[HTML]{fcb379} 71.00 & \cellcolor[HTML]{fcba7b} 67.40 & \cellcolor[HTML]{fcb77a} 69.20 & \cellcolor[HTML]{fdca7e} 62.38 & \cellcolor[HTML]{fdc67d} 59.16 & \cellcolor[HTML]{fdc87d} 60.77\\
 & GTE$_{Base}$ & \cellcolor[HTML]{fcaf79} 30.05 & \cellcolor[HTML]{fed27f} 27.82 & \cellcolor[HTML]{fdbe7b} 28.93 & \cellcolor[HTML]{fed981} 74.57 & \cellcolor[HTML]{fedf82} 70.63 & \cellcolor[HTML]{fedc81} 72.60 & \cellcolor[HTML]{fed780} 64.69 & \cellcolor[HTML]{fed580} 61.76 & \cellcolor[HTML]{fed780} 63.23\\
 & GTE$_{Large}$ & \cellcolor[HTML]{dee182} 40.40 & \cellcolor[HTML]{ffe483} 29.40 & \cellcolor[HTML]{f1e783} 34.90 & \cellcolor[HTML]{fede82} 75.04 & \cellcolor[HTML]{ffe683} 71.23 & \cellcolor[HTML]{ffe282} 73.14 & \cellcolor[HTML]{fede81} 65.78 & \cellcolor[HTML]{fedb81} 62.67 & \cellcolor[HTML]{fedd81} 64.23\\
 & BGE$_{Small}$ & \cellcolor[HTML]{fba677} 29.24 & \cellcolor[HTML]{fdcf7f} 27.49 & \cellcolor[HTML]{fcb87a} 28.37 & \cellcolor[HTML]{fdce7e} 73.49 & \cellcolor[HTML]{fdcc7e} 68.98 & \cellcolor[HTML]{fdcd7e} 71.23 & \cellcolor[HTML]{fed780} 64.59 & \cellcolor[HTML]{fed580} 61.72 & \cellcolor[HTML]{fed780} 63.16\\
 & BGE$_{Base}$ & \cellcolor[HTML]{fb9c75} 28.35 & \cellcolor[HTML]{fdc97d} 27.00 & \cellcolor[HTML]{fcb079} 27.67 & \cellcolor[HTML]{fed17f} 73.83 & \cellcolor[HTML]{fecf7f} 69.23 & \cellcolor[HTML]{fed07f} 71.53 & \cellcolor[HTML]{ffe382} 66.59 & \cellcolor[HTML]{fee182} 63.66 & \cellcolor[HTML]{ffe282} 65.13\\
 & BGE$_{Large}$ & \cellcolor[HTML]{ece583} 38.30 & \cellcolor[HTML]{fed680} 28.14 & \cellcolor[HTML]{feeb84} 33.22 & \cellcolor[HTML]{fedc81} 74.83 & \cellcolor[HTML]{fed981} 70.09 & \cellcolor[HTML]{feda81} 72.46 & \cellcolor[HTML]{ffeb84} 68.05 & \cellcolor[HTML]{ffe683} 64.62 & \cellcolor[HTML]{ffe984} 66.34\\
\midrule
\vl{1em}{12}{\textit{$n=5$}}  & InstructOR$_{Large}$ & \cellcolor[HTML]{d5df82} 41.77 & \cellcolor[HTML]{e6e483} 32.86 & \cellcolor[HTML]{dfe282} 37.31 & \cellcolor[HTML]{e6e483} 78.36 & \cellcolor[HTML]{e9e583} 74.08 & \cellcolor[HTML]{e8e483} 76.22 & \cellcolor[HTML]{e1e282} 70.30 & \cellcolor[HTML]{e5e483} 67.51 & \cellcolor[HTML]{e2e382} 68.90\\
 & E5-v2$_{Base}$ & \cellcolor[HTML]{fee182} 34.49 & \cellcolor[HTML]{fedd81} 28.76 & \cellcolor[HTML]{fedb81} 31.63 & \cellcolor[HTML]{e4e382} 78.53 & \cellcolor[HTML]{efe683} 73.47 & \cellcolor[HTML]{eae583} 76.00 & \cellcolor[HTML]{ffe383} 66.75 & \cellcolor[HTML]{ffe282} 63.94 & \cellcolor[HTML]{ffe383} 65.34\\
 & E5-v2$_{Large}$ & \cellcolor[HTML]{f5e883} 36.82 & \cellcolor[HTML]{ffe583} 29.53 & \cellcolor[HTML]{ffeb84} 33.17 & \cellcolor[HTML]{eae583} 78.02 & \cellcolor[HTML]{ede683} 73.66 & \cellcolor[HTML]{ece683} 75.84 & \cellcolor[HTML]{fedd81} 65.70 & \cellcolor[HTML]{fedb81} 62.76 & \cellcolor[HTML]{fedd81} 64.23\\
 & Multilingual-E5$_{Large}$ & \cellcolor[HTML]{fdbd7b} 31.29 & \cellcolor[HTML]{ffe282} 29.28 & \cellcolor[HTML]{fdcc7e} 30.29 & \cellcolor[HTML]{ffeb84} 76.21 & \cellcolor[HTML]{faea84} 72.18 & \cellcolor[HTML]{fdea84} 74.19 & \cellcolor[HTML]{fed680} 64.36 & \cellcolor[HTML]{fed680} 61.78 & \cellcolor[HTML]{fed680} 63.07\\
 & E5$_{Large}$ & \cellcolor[HTML]{f3e783} 37.24 & \cellcolor[HTML]{e7e483} 32.79 & \cellcolor[HTML]{f0e783} 35.01 & \cellcolor[HTML]{ffe984} 76.04 & \cellcolor[HTML]{ffe683} 71.20 & \cellcolor[HTML]{ffe783} 73.62 & \cellcolor[HTML]{eae583} 69.63 & \cellcolor[HTML]{f0e783} 66.62 & \cellcolor[HTML]{ece683} 68.13\\
 & OpenAI-Ada-002 & \cellcolor[HTML]{c0d980} 45.01 & \cellcolor[HTML]{fed880} 28.38 & \cellcolor[HTML]{e3e382} 36.70 & \cellcolor[HTML]{9ecf7e} 84.56 & \cellcolor[HTML]{9dcf7e} 82.60 & \cellcolor[HTML]{9dcf7e} 83.58 & \cellcolor[HTML]{dae082} 70.81 & \cellcolor[HTML]{dfe282} 68.03 & \cellcolor[HTML]{dce182} 69.42\\
 & GTE$_{Small}$ & \cellcolor[HTML]{fed07f} 32.92 & \cellcolor[HTML]{ffeb84} 30.05 & \cellcolor[HTML]{fed981} 31.48 & \cellcolor[HTML]{fdcb7e} 73.21 & \cellcolor[HTML]{fdce7f} 69.16 & \cellcolor[HTML]{fdcc7e} 71.18 & \cellcolor[HTML]{fedd81} 65.63 & \cellcolor[HTML]{feda81} 62.58 & \cellcolor[HTML]{fedc81} 64.10\\
 & GTE$_{Base}$ & \cellcolor[HTML]{fcae78} 29.90 & \cellcolor[HTML]{ffeb84} 30.02 & \cellcolor[HTML]{fdc97d} 29.96 & \cellcolor[HTML]{fcea84} 76.54 & \cellcolor[HTML]{fbea84} 72.13 & \cellcolor[HTML]{fbea84} 74.33 & \cellcolor[HTML]{ffe683} 67.11 & \cellcolor[HTML]{ffe282} 63.95 & \cellcolor[HTML]{ffe583} 65.53\\
 & GTE$_{Large}$ & \cellcolor[HTML]{d4df82} 41.92 & \cellcolor[HTML]{eae583} 32.41 & \cellcolor[HTML]{e0e282} 37.17 & \cellcolor[HTML]{ffe583} 75.73 & \cellcolor[HTML]{ffe583} 71.18 & \cellcolor[HTML]{ffe583} 73.45 & \cellcolor[HTML]{f9e984} 68.48 & \cellcolor[HTML]{ffeb84} 65.38 & \cellcolor[HTML]{fbea84} 66.93\\
 & BGE$_{Small}$ & \cellcolor[HTML]{ffeb84} 35.33 & \cellcolor[HTML]{e8e483} 32.64 & \cellcolor[HTML]{f8e984} 33.99 & \cellcolor[HTML]{fdc77d} 72.85 & \cellcolor[HTML]{fdc27c} 68.06 & \cellcolor[HTML]{fdc47d} 70.46 & \cellcolor[HTML]{ffe683} 67.15 & \cellcolor[HTML]{ffe583} 64.35 & \cellcolor[HTML]{ffe683} 65.75\\
 & BGE$_{Base}$ & \cellcolor[HTML]{fa9774} 27.94 & \cellcolor[HTML]{ffe583} 29.49 & \cellcolor[HTML]{fcbb7b} 28.72 & \cellcolor[HTML]{fbea84} 76.61 & \cellcolor[HTML]{fdea84} 71.90 & \cellcolor[HTML]{fcea84} 74.25 & \cellcolor[HTML]{ede683} 69.42 & \cellcolor[HTML]{f1e783} 66.52 & \cellcolor[HTML]{eee683} 67.97\\
 & BGE$_{Large}$ & \cellcolor[HTML]{fcea84} 35.79 & \cellcolor[HTML]{eae583} 32.38 & \cellcolor[HTML]{f8e984} 34.08 & \cellcolor[HTML]{ffeb84} 76.26 & \cellcolor[HTML]{ffe383} 71.00 & \cellcolor[HTML]{ffe783} 73.63 & \cellcolor[HTML]{dce182} 70.68 & \cellcolor[HTML]{e4e382} 67.64 & \cellcolor[HTML]{dfe282} 69.16\\
\midrule
\vl{1em}{12}{\textit{$n=10$}}  & InstructOR$_{Large}$ & \cellcolor[HTML]{b0d47f} 47.38 & \cellcolor[HTML]{dee182} 33.77 & \cellcolor[HTML]{c6da81} 40.58 & \cellcolor[HTML]{ccdc81} 80.58 & \cellcolor[HTML]{d4de81} 76.50 & \cellcolor[HTML]{d1de81} 78.54 & \cellcolor[HTML]{c5da81} 72.37 & \cellcolor[HTML]{cbdc81} 69.68 & \cellcolor[HTML]{c8db81} 71.03\\
 & E5-v2$_{Base}$ & \cellcolor[HTML]{f4e883} 37.04 & \cellcolor[HTML]{ece683} 32.17 & \cellcolor[HTML]{f4e883} 34.60 & \cellcolor[HTML]{cfdd81} 80.31 & \cellcolor[HTML]{e2e382} 74.92 & \cellcolor[HTML]{dae082} 77.61 & \cellcolor[HTML]{eae583} 69.59 & \cellcolor[HTML]{ede683} 66.86 & \cellcolor[HTML]{ebe583} 68.23\\
 & E5-v2$_{Large}$ & \cellcolor[HTML]{b4d580} 46.80 & \cellcolor[HTML]{e9e583} 32.53 & \cellcolor[HTML]{cddc81} 39.66 & \cellcolor[HTML]{dee182} 79.11 & \cellcolor[HTML]{e7e483} 74.31 & \cellcolor[HTML]{e3e382} 76.71 & \cellcolor[HTML]{f7e984} 68.65 & \cellcolor[HTML]{fbea84} 65.70 & \cellcolor[HTML]{f8e984} 67.17\\
 & Multilingual-E5$_{Large}$ & \cellcolor[HTML]{fcb97a} 30.88 & \cellcolor[HTML]{e7e483} 32.70 & \cellcolor[HTML]{fedd81} 31.79 & \cellcolor[HTML]{e2e382} 78.71 & \cellcolor[HTML]{e6e483} 74.43 & \cellcolor[HTML]{e5e382} 76.57 & \cellcolor[HTML]{ffea84} 67.87 & \cellcolor[HTML]{ffeb84} 65.39 & \cellcolor[HTML]{ffeb84} 66.63\\
 & E5$_{Large}$ & \cellcolor[HTML]{d7e082} 41.44 & \cellcolor[HTML]{d5df82} 34.74 & \cellcolor[HTML]{d9e082} 38.09 & \cellcolor[HTML]{ede683} 77.83 & \cellcolor[HTML]{f0e783} 73.35 & \cellcolor[HTML]{efe683} 75.59 & \cellcolor[HTML]{c4da81} 72.42 & \cellcolor[HTML]{ccdc81} 69.62 & \cellcolor[HTML]{c8db81} 71.02\\
 & OpenAI-Ada-002 & \cellcolor[HTML]{b6d680} 46.60 & \cellcolor[HTML]{e6e483} 32.90 & \cellcolor[HTML]{ccdc81} 39.75 & \cellcolor[HTML]{92cc7e} 85.57 & \cellcolor[HTML]{95cc7e} 83.46 & \cellcolor[HTML]{94cc7e} 84.51 & \cellcolor[HTML]{b8d780} 73.30 & \cellcolor[HTML]{c0d980} 70.60 & \cellcolor[HTML]{bcd880} 71.95\\
 & GTE$_{Small}$ & \cellcolor[HTML]{fdcd7e} 32.71 & \cellcolor[HTML]{e0e282} 33.53 & \cellcolor[HTML]{ffeb84} 33.12 & \cellcolor[HTML]{fedb81} 74.77 & \cellcolor[HTML]{fedd81} 70.42 & \cellcolor[HTML]{fedc81} 72.59 & \cellcolor[HTML]{ffe883} 67.48 & \cellcolor[HTML]{ffe683} 64.56 & \cellcolor[HTML]{ffe783} 66.02\\
 & GTE$_{Base}$ & \cellcolor[HTML]{fb9974} 28.05 & \cellcolor[HTML]{f4e883} 31.23 & \cellcolor[HTML]{fdc57d} 29.64 & \cellcolor[HTML]{f2e783} 77.35 & \cellcolor[HTML]{f5e883} 72.76 & \cellcolor[HTML]{f4e883} 75.06 & \cellcolor[HTML]{ebe583} 69.50 & \cellcolor[HTML]{f2e783} 66.44 & \cellcolor[HTML]{eee683} 67.97\\
 & GTE$_{Large}$ & \cellcolor[HTML]{c0d980} 45.05 & \cellcolor[HTML]{d1de81} 35.25 & \cellcolor[HTML]{c9db81} 40.15 & \cellcolor[HTML]{ffeb84} 76.29 & \cellcolor[HTML]{ffeb84} 71.67 & \cellcolor[HTML]{ffeb84} 73.98 & \cellcolor[HTML]{e7e483} 69.86 & \cellcolor[HTML]{ede683} 66.90 & \cellcolor[HTML]{e9e583} 68.38\\
 & BGE$_{Small}$ & \cellcolor[HTML]{f9e984} 36.24 & \cellcolor[HTML]{d8e082} 34.44 & \cellcolor[HTML]{eee683} 35.34 & \cellcolor[HTML]{ffe883} 75.95 & \cellcolor[HTML]{ffe583} 71.13 & \cellcolor[HTML]{ffe683} 73.54 & \cellcolor[HTML]{f3e783} 68.96 & \cellcolor[HTML]{f4e883} 66.27 & \cellcolor[HTML]{f3e783} 67.61\\
 & BGE$_{Base}$ & \cellcolor[HTML]{fcbc7b} 31.14 & \cellcolor[HTML]{f1e783} 31.62 & \cellcolor[HTML]{fed880} 31.38 & \cellcolor[HTML]{e9e583} 78.15 & \cellcolor[HTML]{f2e783} 73.07 & \cellcolor[HTML]{eee683} 75.61 & \cellcolor[HTML]{d1de81} 71.48 & \cellcolor[HTML]{d7df82} 68.73 & \cellcolor[HTML]{d3de81} 70.10\\
 & BGE$_{Large}$ & \cellcolor[HTML]{ccdc81} 43.19 & \cellcolor[HTML]{cedd81} 35.56 & \cellcolor[HTML]{cfdd81} 39.38 & \cellcolor[HTML]{ede683} 77.77 & \cellcolor[HTML]{f8e984} 72.44 & \cellcolor[HTML]{f4e883} 75.10 & \cellcolor[HTML]{c5da81} 72.36 & \cellcolor[HTML]{cfdd81} 69.39 & \cellcolor[HTML]{c9dc81} 70.88\\
\midrule
\vl{1em}{12}{\textit{$n=15$}}  & InstructOR$_{Large}$ & \cellcolor[HTML]{dde182} 40.59 & \cellcolor[HTML]{cfdd81} 35.40 & \cellcolor[HTML]{dae082} 37.99 & \cellcolor[HTML]{ccdc81} 80.57 & \cellcolor[HTML]{dae082} 75.75 & \cellcolor[HTML]{d4df82} 78.16 & \cellcolor[HTML]{bbd780} 73.10 & \cellcolor[HTML]{c1d980} 70.54 & \cellcolor[HTML]{bed880} 71.82\\
 & E5-v2$_{Base}$ & \cellcolor[HTML]{d2de81} 42.17 & \cellcolor[HTML]{d8e082} 34.44 & \cellcolor[HTML]{d7df82} 38.31 & \cellcolor[HTML]{d0dd81} 80.25 & \cellcolor[HTML]{e4e382} 74.65 & \cellcolor[HTML]{dce182} 77.45 & \cellcolor[HTML]{e2e382} 70.18 & \cellcolor[HTML]{e6e483} 67.50 & \cellcolor[HTML]{e3e382} 68.84\\
 & E5-v2$_{Large}$ & \cellcolor[HTML]{aed47f} 47.71 & \cellcolor[HTML]{dfe282} 33.67 & \cellcolor[HTML]{c5da81} 40.69 & \cellcolor[HTML]{d5df82} 79.86 & \cellcolor[HTML]{e4e382} 74.66 & \cellcolor[HTML]{dee182} 77.26 & \cellcolor[HTML]{e9e583} 69.70 & \cellcolor[HTML]{efe683} 66.69 & \cellcolor[HTML]{ebe583} 68.19\\
 & Multilingual-E5$_{Large}$ & \cellcolor[HTML]{fb9c75} 28.31 & \cellcolor[HTML]{e0e282} 33.48 & \cellcolor[HTML]{fed37f} 30.89 & \cellcolor[HTML]{d4df82} 79.91 & \cellcolor[HTML]{dee282} 75.32 & \cellcolor[HTML]{dae082} 77.61 & \cellcolor[HTML]{eee683} 69.31 & \cellcolor[HTML]{eee683} 66.76 & \cellcolor[HTML]{ede683} 68.03\\
 & E5$_{Large}$ & \cellcolor[HTML]{d1de81} 42.42 & \cellcolor[HTML]{c7db81} 36.31 & \cellcolor[HTML]{cfdd81} 39.36 & \cellcolor[HTML]{eae583} 78.02 & \cellcolor[HTML]{f3e883} 73.00 & \cellcolor[HTML]{efe783} 75.51 & \cellcolor[HTML]{bbd780} 73.13 & \cellcolor[HTML]{c4da81} 70.26 & \cellcolor[HTML]{bfd980} 71.69\\
 & OpenAI-Ada-002 & \cellcolor[HTML]{acd37f} 48.13 & \cellcolor[HTML]{d9e082} 34.26 & \cellcolor[HTML]{c1d980} 41.20 & \cellcolor[HTML]{81c77d} 87.04 & \cellcolor[HTML]{87c87d} 85.03 & \cellcolor[HTML]{84c87d} 86.03 & \cellcolor[HTML]{afd47f} 73.97 & \cellcolor[HTML]{b7d680} 71.36 & \cellcolor[HTML]{b3d580} 72.66\\
 & GTE$_{Small}$ & \cellcolor[HTML]{eae583} 38.54 & \cellcolor[HTML]{d8e082} 34.38 & \cellcolor[HTML]{e5e483} 36.46 & \cellcolor[HTML]{fede82} 75.03 & \cellcolor[HTML]{fedc81} 70.32 & \cellcolor[HTML]{fedd81} 72.68 & \cellcolor[HTML]{f7e984} 68.63 & \cellcolor[HTML]{fcea84} 65.60 & \cellcolor[HTML]{f9e984} 67.12\\
 & GTE$_{Base}$ & \cellcolor[HTML]{fed880} 33.68 & \cellcolor[HTML]{eae583} 32.35 & \cellcolor[HTML]{ffea84} 33.02 & \cellcolor[HTML]{e7e483} 78.27 & \cellcolor[HTML]{eee683} 73.56 & \cellcolor[HTML]{ebe583} 75.92 & \cellcolor[HTML]{e7e483} 69.86 & \cellcolor[HTML]{efe683} 66.73 & \cellcolor[HTML]{eae583} 68.29\\
 & GTE$_{Large}$ & \cellcolor[HTML]{eee683} 37.98 & \cellcolor[HTML]{d8e082} 34.38 & \cellcolor[HTML]{e7e483} 36.18 & \cellcolor[HTML]{ede683} 77.78 & \cellcolor[HTML]{f4e883} 72.93 & \cellcolor[HTML]{f1e783} 75.36 & \cellcolor[HTML]{dee182} 70.51 & \cellcolor[HTML]{e4e382} 67.62 & \cellcolor[HTML]{e0e282} 69.07\\
 & BGE$_{Small}$ & \cellcolor[HTML]{fb9974} 28.06 & \cellcolor[HTML]{d9e082} 34.30 & \cellcolor[HTML]{fed680} 31.18 & \cellcolor[HTML]{ffe282} 75.43 & \cellcolor[HTML]{fede82} 70.54 & \cellcolor[HTML]{fee082} 72.98 & \cellcolor[HTML]{e2e382} 70.20 & \cellcolor[HTML]{e5e382} 67.56 & \cellcolor[HTML]{e3e382} 68.88\\
 & BGE$_{Base}$ & \cellcolor[HTML]{fa8f72} 27.20 & \cellcolor[HTML]{f6e883} 31.08 & \cellcolor[HTML]{fdc07c} 29.14 & \cellcolor[HTML]{e0e282} 78.92 & \cellcolor[HTML]{ede683} 73.65 & \cellcolor[HTML]{e7e483} 76.29 & \cellcolor[HTML]{cbdc81} 71.93 & \cellcolor[HTML]{d2de81} 69.15 & \cellcolor[HTML]{cedd81} 70.54\\
 & BGE$_{Large}$ & \cellcolor[HTML]{d2de81} 42.22 & \cellcolor[HTML]{c0d980} 37.06 & \cellcolor[HTML]{cddd81} 39.64 & \cellcolor[HTML]{e2e382} 78.76 & \cellcolor[HTML]{efe683} 73.43 & \cellcolor[HTML]{e9e583} 76.10 & \cellcolor[HTML]{bad780} 73.17 & \cellcolor[HTML]{c5da81} 70.24 & \cellcolor[HTML]{bfd980} 71.71\\
 \bottomrule
    \end{tabular}
    \caption{Results per model using $k$ synthetic examples averaged across 20 samples.}
    \label{tab:appendix:synthetic-full}
\end{table*}

\section{Table of averaged mean and standard deviation statistics for examples generated using gpt-4}
\label{sec:appendix:synthetic-gpt4}
See Table \ref{tab:appendix:synthetic-examples-gpt4}.

\begin{table*}
    \centering
    \small
    \setlength{\tabcolsep}{3pt}
    \begin{tabular}{c| >{\normalsize} l|c|c|c|c|c|c}
    \toprule
    \multirow{2}{*}{$k$}& \multirow{2}{*}{\normalsize \textbf{Metric}} & \multicolumn{2}{c |}{\normalsize\textbf{ATIS}} & \multicolumn{2}{c |}{\normalsize\textbf{SNIPS}} & \multicolumn{2}{c }{\normalsize\textbf{CLINC}} \\
    \cline{3-8}
    & & \normalsize $\mu$ & \normalsize $\sigma$ & \normalsize $\mu$ & \normalsize $\sigma$ & \normalsize $\mu$ & \normalsize $\sigma$ \\
    \midrule
\vl{1em}{3}{\textit{$k=1$}}  & Mean & 24.51 & 10.15 & 67.63 & 5.48 & 51.63 & 5.13\\
 & $\Delta_{Label}$ & -7.19 & -2.58 & -8.68 & -1.05 & -15.56 & 0.08\\
 & $\Delta_{Desc}$ & -27.38 & 6.37 & -19.29 & 2.46 & -22.92 & 2.12\\
\midrule
\vl{1em}{3}{\textit{$k=3$}}  & Mean & 31.19 & 8.61 & 73.25 & 4.49 & 63.71 & 2.76\\
 & $\Delta_{Label}$ & -0.51 & -4.11 & -3.06 & -2.04 & -3.47 & -2.29\\
 & $\Delta_{Desc}$ & -20.70 & 4.84 & -13.66 & 1.47 & -10.83 & -0.25\\
\midrule
\vl{1em}{3}{\textit{$k=5$}}  & Mean & 33.29 & 7.90 & 74.73 & 4.16 & 66.54 & 2.35\\
 & $\Delta_{Label}$ & 1.59 & -4.82 & -1.57 & -2.37 & -0.64 & -2.70\\
 & $\Delta_{Desc}$ & -18.60 & 4.13 & -12.18 & 1.14 & -8.00 & -0.67\\
\midrule
\vl{1em}{3}{\textit{$k=10$}}  & Mean & 36.12 & 7.51 & 76.28 & 3.49 & 68.92 & 2.08\\
 & $\Delta_{Label}$ & 4.42 & -5.21 & -0.02 & -3.04 & 1.73 & -2.97\\
 & $\Delta_{Desc}$ & -15.77 & 3.73 & -10.63 & 0.48 & -5.63 & -0.94\\
\midrule
\vl{1em}{3}{\textit{$k=15$}}  & Mean & 36.17 & 7.13 & 76.78 & 3.75 & 69.74 & 1.93\\
 & $\Delta_{Label}$ & 4.47 & -5.59 & 0.48 & -2.78 & 2.55 & -3.12\\
 & $\Delta_{Desc}$ & -15.72 & 3.36 & -10.13 & 0.73 & -4.81 & -1.09\\
    \bottomrule
    \end{tabular}
    \caption{Averaged mean of accuracy and macro-f1 scores experiments conducted across 20 samples and 12 models using $k$ number of synthetic examples per intent class generated using \texttt{gpt-4-1106-preview}. $\Delta_{Label}$ and $\Delta_{Desc}$ are differences to the averaged performance of methods using tokenized labels and intent descriptions respectively.}
    \label{tab:appendix:synthetic-examples-gpt4}
\end{table*}

\end{document}